\documentclass[twoside,11pt]{article}

\usepackage{blindtext}

\usepackage[preprint]{jmlr2e}

\usepackage{jmlr2e}

\usepackage{lastpage}
\usepackage[margin=1in]{geometry}
\usepackage{amssymb}
\usepackage{upgreek}
\usepackage{algorithm}
\usepackage{algpseudocode}
\usepackage{graphicx}
\usepackage{array}
\usepackage{amsmath}
\usepackage{makecell}
\usepackage{multirow}
\usepackage{enumitem}
\usepackage{caption}

\ShortHeadings{Covariance-Boosted Gaussian Processes}{Ovadia}
\firstpageno{1}

\begin{document}

\title{Covariance-Boosted Gaussian Processes for Spatiotemporal Irregularities}

\author{\name Jeremy Ovadia \email jeremy@sequoiarc.com \\
       \addr Sequoia Research Corporation\\
       2875 Temple Ave\\
       Signal Hill, CA, 90755, USA}

\editor{Insert editor here.}

\maketitle


\begin{abstract}
Nonstationary Gaussian process (GP) models are powerful tools for capturing input-dependent variability by adapting to observed data. However, with limited sampling and highly parameterized covariance structure, they are often prone to overfitting and overconfident uncertainty estimates, potentially leading to misleading predictions in safety-critical applications. Motivated by ionospheric modeling for satellite-based augmentation systems (SBAS), this paper proposes a Covariance-Boosted Gaussian Process (CBGP) framework centered upon boosting covariance priors to discover nonstationary latent functions for signal and observation variation that capture irregularities in the input domain. An additional layer of GP modeling of ``partially-whitened'' observations guides latent function relative error estimation that is used to iteratively update weak priors in a gradient descent-like procedure. Following boosting, restrictions are imposed upon prior covariances to prevent overfitting while posterior uncertainties are inflated to prevent model overconfidence. CBGP model efficacy and robustness are demonstrated through out-of-sample testing of both simulated and real-world applications that meet a three-nines integrity standard. The modeling of an extensive ionospheric storm dataset over South America suggests accurate and reliable means to compute SBAS ionospheric corrections in the most challenging space weather environment using regional models that are more informed and responsive than local fitting performed by currently-operating SBAS.
\end{abstract}



\section{Introduction}
\label{sec: introduction}

As machine learning and artificial intelligence models have become more widely adopted, they have been increasingly considered for deployment in safety-critical applications. Specifically, continual real-time probabilistic inference models have been employed for monitoring, decision making, and fault detection in integrity-demanding settings that include autonomous driving \citep{viveiros2025}, intensive care units in hospitals \citep{shillan2019}, nuclear reactors \citep{yu2020}, and air traffic control \citep{shone2021} among several others. Such models must be reliable, robust, and computationally-feasible to be executed in real-time settings, balancing tradeoffs of accuracy, efficiency, and consistency. Ultimately, applications to safety-critical systems cannot be overly forgiving from a safety perspective and often have integrity requirements that constrain models.

The integrity of continual real-time probabilistic inference models is most severely challenged under irregular operating conditions or when anomalous observations are encountered, where the risk of producing hazardously misleading information (HMI) is elevated. The mishandling of irregularities by inference models may yield misleading information (MI) that can have significant repercussions for integrity-focused applications, such as financial risk management \citep{liu2020b} and aerospace structure monitoring \citep{alvarezmontoya2019}. However, once MI poses risk to human well-being, as may be the case with applications to medical imaging \citep{dahnke2014} and natural disaster risk assessment \citep{wang2025}, does it become HMI. Under these conditions, proper treatment of irregularities through model robustness and integrity become the highest priorities to prevent HMI. In safety-critical contexts, inference models designed with perfomance objectives to optimize accuracy under nominal working conditions, or even disturbed conditions, may not provide necessary safety that human life relies on.

In both safety-critical and non-safety-critical applications, Gaussian processes (GPs) have widespread utility in describing real-world complex phenomenon that appear in various natural and computational contexts. A GP, expressed simply, is a stochastic process, or a collection of random variables, that can be jointly described by Gaussian distributions. Gaussian process regression (GPR) is a non-parametric, Bayesian interpolation technique that optimally determines linear weights of sampled observations as a best linear unbiased predictor (BLUP). GP modeling, generally, provides posterior expected values and uncertainties that describe predictive Gaussian distributions at unsampled states.

GP models can be tuned, account for heteroskedastic noise, or include nonstationarity as means to better handle irregularities that pose a threat to a system being modeled; however, the ability for these models to adapt to input observations may also be accompanied by fragility that increases the risks of not meeting potential integrity standards, or producing HMI in a safety-critical setting. As adept as they may be in describing observations, nonstationary GP models may expose practitioners to overfitting and overconfidence, particularly with many hyperparameters and small samplings \citep{mohammed2017}. Covariance tuning via leave-one-out cross-validation have been shown to be susceptible to overfitting due to relatively high variances \citep{cawley2007}. Simulation-based methods that account for heteroskedastic noise have been often cited as potentially unreliable in scenarios with small sample numbers or high noise levels \citep{zhang2016, wang2019}. The use of priors and regularization are two common remedies for overfitting with GP models \citep[preface]{rasmussen2006} but may not always be easily-integrated into advanced adaptive models.

A notable GPR application to a safety-critical system subject to HMI risk is ionospheric modeling for satellite-based augmentation systems (SBAS). The Wide Area Augmentation System (WAAS), the Federal Aviation Association's augmentation of the Global Positioning System (GPS), broadcasts SBAS corrections to aviation users for safe runway approaches across the United States, which include corrections for ionospheric delays, the largest source of user positioning error \citep{milliken1978}. WAAS' ionospheric corrections, including gridded delay estimates and corresponding uncertainties, are computed in real-time using a stationary GP model that relies on a few accomodations to account for the spatial inhomogeneity of the ionosphere. First, WAAS omits all measurements near the geomagnetic equator, where the ionosphere is more active than at mid-latitudes, using a pierce point filter line \citep{sparks2022}.  Next, ionospheric models sub-sample measurements around local neighborhoods around each point in a set of spatial ionospheric grid points (IGPs) to compute corrections to be interpolated by the user \citep{sparks2011a}. Finally, WAAS' uncertainties for ionospheric corrections are inflated post-fit as means to mitigate the risk of broadcasting HMI \citep{walter2001, sparks2011b}. To further reduce the risk of broadcasting HMI from undersampling, WAAS also protects its users from potential spatiotemporal variation of the ionosphere that is extremely challenging to predict with limited observations using an ``ionospheric threat model'' \citep{blanch2002}. The result is robust, efficient homogeneous ionospheric modeling performed in real-time that has successfully aided aerial guidance over the United States for decades but, ultimately, is a local stationary model with limited adaptability.

In this paper, a nonstationary GP modeling framework based on boosting nonstationary covariance priors using GP-based relative error approximations is developed and tested. Following a brief review of GP fundamentals and related work, approximations for these relative errors are modeled in the domain of observations whitened using prior covariance approximations. Then, an integrity-focused boosting algorithm that includes covariance inflation of both priors and posteriors is proposed. Aspects of the modeling framework are illustrated by a simulated application problem for which an irregularity is injected into a stationary GP. Next, the framework is tested on out-of-sample predictions for motorcycle acceleration \citep{silverman1985} and Meuse River \citep{rikken1993} benchmark problems against other existing heterogeneous GP models. Lastly, a vast data set of tens of millions ionospheric measurements over South America is modeled with a focus on both accuracy and integrity of predictions to demonstrate the usefulness of our models for SBAS.


\section{Background and Related Work}
\label{sec: background}

Several aspects of GP modeling, including adaptivity, misspecification, and error bounding, have been well-studied and are worth discussing before outlining novel GP methods. Fundamental concepts, foundational work, and relevant advancements are presented in this section to provide the groundwork for models presented in Section \ref{sec: models}.


\subsection{Gaussian Process Formulation and Regression}
\label{subsec: gaussian_process}

Consider a stochastic process $y_x \in \mathbb{R}$ continuous with respect to a spatiotemporal quantity $x\in  \mathbb{R}^n$. Assume that this process is a GP that can be fully-described by its mean, $\mu(x)=\mathbb{E}\left[y_x\right]$, and covariance function, $C_y(x, x') = \mathbb{E}[(y_x - \mu(x))(y_{x'} - \mu(x'))]$. Consider a finite sampling of this process $\{(x_i, y_i)\}_{i=1}^N$, and let $C$ denote the corresponding covariance matrix with entries $C_{i,j} = C_y(x_i, x_j)$. Suppose that the kernel function for the covariance matrix takes the following form:
\begin{equation}
\label{eq:kernel}
k\left(u, v\right) = \sigma_{\mathrm{signal}}\left(u\right)\sigma_{\mathrm{signal}}\left(v\right)\rho\left(u, v\right)
\end{equation}
where $ \sigma_{\mathrm{signal}}$ denotes the variation of the signal and $\rho$ describes its correlation function. The kernel evaluated at the sampled states can be written in matrix form as
\begin{equation}
\label{eq:kernel_matrix}
K=\Sigma\uprho\Sigma
\end{equation}
where $K$ denotes the matrix computed by the kernel function with $K_{i,j}=k\left(x_i, x_j\right)$, $\uprho$ is a correlation matrix with $\uprho_{i,j} = \rho\left(x_i, x_j\right)$, and $\Sigma$ is a diagonal matrix with $\Sigma_{i,i} =  \sigma_{\mathrm{signal}}\left(x_i\right)$. Further assume that $\uprho$ is symmetric positive definite (SPD), implying $C$ is an SPD matrix as well.

The two commonly-used positive definite correlation functions of primary focus in this study are an exponential decay, or Ornstein-Uhlenbeck, correlation function
\begin{equation}
\label{eq: exponential_decay_kernel}
\rho_\mathrm{OU}\left(x, x^\prime\right) = \exp\left(-\frac{\lVert x - x^\prime \rVert}{L}\right)
\end{equation}
and the squared exponential decay, or Gaussian radial basis function (RBF), correlation function
\begin{equation}
\label{eq: exponential_decay_kernel}
\rho_\mathrm{RBF}\left(x, x^\prime\right) = \exp\left(-\frac{\lVert x - x^\prime \rVert^2}{L^2}\right).
\end{equation}
Alternative representations for each of these correlation functions may include a $\frac{1}{2}$ scaling factor within their exponential arguments. Both correlation functions are positive definite.

Let $\sigma^2_\mathrm{obs}\left(u\right)$ denote the observation variance of the sampling at the spatiotemporal state $u$, which is assumed to include potential measurement noise and nugget effects. Let $R$ be the diagonal observation variance matrix sampled at $\{x_i\}$ defined by $R_{i,i} = \sigma^2_\mathrm{obs}\left(x_i\right)$. Then, the prior covariance for the underlying GP that accounts for observation variance can be expressed as
\begin{equation}
\label{eq:C_equal_K_plus_R}
C = \Sigma\uprho\Sigma + R.
\end{equation}

This prior covariance guides the whitening of the vector of observations, $y = \left(y_i\right)$, using a whitening matrix $\Gamma$ satisfying
\begin{equation}
\label{eq:whitening_definition}
\Gamma^T\Gamma=C^{-1}
\end{equation}
 where
\begin{equation}
\label{eq:whitening_sim}
\Gamma y \sim \mathcal{N}\left(0,I\right)
\end{equation}
takes the form of unit-normal white noise.
A specific form for the whitening matrix $\Gamma$ can be uniquely determined using various methods, such as Mahalanobis (or Zero Components Analysis) whitening, Principal Components Analysis whitening, or Cholesky decomposition. This study will focus on Mahalanobis whitening by the matrix square root of the inverse of the covariance, represented by
\begin{equation}
\label{eq:mahalanobis}
\Gamma = C^{-1/2},
\end{equation}
which is well-defined for an SPD covariance matrix.

A GPR model interpolates the sampling of a GP $\{(x_i, y_i)\}_{i=1}^N$ at any vector of spatiotemporal states, $x^*$, whether sampled or unsampled, to predicted observable states, $y^*$. Similar to GPs, GPR models are described solely by their posterior means and covariances, which are given as follows:
\begin{equation}
\label{eq: gpr_mean}
\mathbb{E}\left[ y^* \right] = w^* y
\end{equation}
\begin{equation}
\label{eq: gpr_cov}
\mathrm{cov}\left[ y^* \right] = K^{**} + R^* -  w^* K^{*T}
\end{equation}
where
\begin{equation}
\label{eq: w_def}
w^* =K^* C^{-1}.
\end{equation}
Here, $K_{i,j}^* = k(x_i, x_j^*)$ and $K_{i,j}^{**} = k(x_i^*, x_j^*)$ define the kernel matrices $K^*$ and $K^{**}$, respectively. $R^*$ is the observation variance matrix at $x^*$ specified by $R_{i,i}^*=\sigma^2_\mathrm{obs}\left(x_i^*\right)$. It is important to note that the posterior covariance for the GPR solution in Equation \eqref{eq: gpr_cov} is independent of the actual observations, $y$, but, instead, just the prior covariance and its components.

The gradient of the expected value of the GPR solution with respect to $x^*$  is given by
\begin{equation}
\label{eq: gradient_mean}
\nabla\left(\mathbb{E}\left[ y^* \right]\right) = \nabla\left(K^*\right) C^{-1}y
\end{equation}
where $ \nabla\left(K^*\right)$ can be evaluated at $x^*$ using a prescribed or inferred functional form for the kernel function $k$.


\subsection{Related Advancements In Gaussian Process Modeling}
\label{subsec: existing_heterogeneous_gp}

The kernels and observation noise, or nugget component, that comprise the GP prior covariance are critical aspects in describing similarity of observations and have been primary areas of focus in the development of GPR models \citep[chap.~4]{rasmussen2006}. Popular approaches for the tuning of hyperparameters defining these elements include optimizing marginal likelihoods and leave-one-out cross validation \citep[chap.~5]{rasmussen2006}. Other kernel adaption methods involve semidefinite programming problem for optimal Gram matrix alignment \citep{lanckriet2004} and hybrid kernels with genetic algorithm determined gain coefficients \citep{ouyang2023}. 

Kernels may also be discovered and tuned to observations through constructive methods. Additive kernel construction methods adds simple kernels together to yield more complex ones \citep[chap.~4]{rasmussen2006}. Iterative kernel discovery, on the other hand, may utilize additive and multiplicative components for structural kernel expansion in an adaptive multi-step fashion \citep{lloyd2014}. Other approaches have combined GPs with gradient boosting decision trees to enhance the ability to account for nonlinear feature interactions \citep{sigrist2022}.

The tuning and adaptivity of stationary GPs with kernels and noise properties that are assumed to be uniform throughout the input domain may better model observations but will fail to properly describe heterogeneous conditions that necessitate inhomogeneity of the prior covariance. Heteroskedastic GPs model observation noise nonuniformly and assume it depends on the observation state, meaning that the observation variance matrix $R$ is not a scalar multiple of the identity matrix. Smooth heteroskedastic noise models using Markov chain Monte Carlo methods can estimate the posterior noise variance \citep{goldberg1998} with extensions including convex optimization  \citep{le2005} and secondary GP models for noise \citep{kersting2007}. Stochastic Kriging serves as a simulation-based metamodeling tool that decouples deterministic trends from heteroskedastic noise components \citep{ankenman2008}. Recent software packages leveraging modern heteroskedastic GP modeling approaches include GPflow \citep{gpflow2017}, BoTorch \citep{balandat2020}, PyMC \citep{pymc2023}, and hetGP \citep{binois2018b}. The software package hetGPy \citep{ogara2025}, which is a Python implementation of hetGP, will be included in computational experiments in Section \ref{subsec: benchmark_experiments}.

Beyond heteroskedastic GPs, nonstationary GP models provide means to describe intricate relationships that may vary significantly within the fit domain. Utilizing nonstationary kernels with signal variances and length scales that are inhomogeneous allows GPs to account for dynamics that are not solely dependent upon relative distances between observations. As is the case when accounting for heteroskedastic noise, nonstationary models must attribute the source of inhomogeneity of observations to signal and observation noise components; however, in doing so, nonstationary models employ methods to adapt both signal and noise components to realized observations using various methods \citep{booth2023, noack2024}. Estimating these kernel components as smooth latent functions embedded within the model introduces more hyperparameters but can provide significant more explanatory power in capturing inhomogeneity in GP modeling \citep{plagemann2008, heinonen2016}.

A common approach to model nonstationarity within a GP is to account for an inhomogeneous scaling of the input domain within the kernel. Input warping accounts for nonstationarity by transforming the input domain with expansion in areas of high variation and compression in areas of low variation \citep{snoek2014}. Another input scaling approach is to explicitly model kernel length scales using inhomogeneous functions. One such kernel in a Gaussian RBF form was proposed in one-dimension by \citet{gibbs1997} and later generalized to multiple dimensions with anisotropic local geometry by \citet{paciorek2003}. In computational experiments in Section \ref{subsec: benchmark_experiments}, an isotropic Gibbs-Paciorek kernel in $N_d$ dimensions given by
\begin{equation}
\label{eq: gibbs_paciorek}
k_\ell(x,x')
=
\sigma_\mathrm{signal}(x)\sigma_\mathrm{signal}(x')
\left(
\frac{2\,\ell(x)\,\ell(x')}
{\ell(x)^2+\ell(x')^2}
\right)^{N_d/2}
\exp\!\left(
-\frac{\|x-x'\|^2}
{\ell(x)^2+\ell(x')^2}
\right),
\end{equation}
where $\ell$ is modeled as a latent function parameterized by an expansion of Gaussian RBF bases is tested against other existing and novel methods discussed in this paper.

Other nonstationary methods account for inhomogeneity without explicit functional forms guiding a smooth nonstationary kernel. Divide-and-conquer methods \citep{rasmussen2001, kim2005} and treed GPs \citep{gramacy2008} partition the input space and leverage multiple stationary GPs in a nonstationary fashion. Nonstationary GPs have also been constructed using Fourier components with frequencies inferred from using neural networks \citep{ton2018}. Deep Gaussian processes use successive GP layers to construct nonstationary kernels learned from training data \citep{damianou2013, desouza2023} while deeply nonstationary GPs layer nonstationary GPs in such a way that length a single layer informs the length scale of the following one \citep{salimbeni2017}.


\subsection{Model Assessment and Error Bounding}
\label{subsec: model_assessment_and_error_bounding}

Due to challenges in accurately describing real-world observations using a GP, model misspecification and criticism for GPs has been a key area of focus for practitioners. When a GP is misspecified, meaning it misaligns with the true process or system, posterior mean and variance estimates may be inaccurate and provide MI. These inaccuracies often manifest in the form of model overconfidence when misspecification results from the presence of irregularities or disturbances to the system. Adaptive GP methods that are responsive to input observations, such as those mentioned in Section \ref{subsec: existing_heterogeneous_gp}, may be misguided and become excessively overconfident in undersampled contexts. Studying predictive simulations and standard residuals are straightforward means to assess GP posteriors. However, residual diagnostics in the whitened domain, as specified by Equation \eqref{eq:whitening_sim}, can provide an enhanced insight into kernel misspecification as opposed to just identifying individual observations that are poorly fit by GPR posteriors \citep{bastos2009}.

To meet potential integrity standards, posterior uncertainty estimates for model predictions can be inflated after fits are performed to account for either model misspecification or the undersampling of observations used for fitting an underlying process. A simple example of this would be inflating an empirical variance estimate using the corresponding standard error for the variance for a population when the independent sample number, $N$, is relatively low. GP upper confidence bound (UCB) models effectively inflate uncertainty by inflating posterior variances by $\beta_t$ to guide future sampling of a GP \citep{srinivas2010}. Other UCB-like extensions take a rigorous approach to specifying the inflation factor $\beta$ for safe control applications \citep{lederer2019, fiedler2021}. 

One approach for posterior uncertainty inflation, referred to here as $R_\mathrm{irreg}$ inflation, is an existing spatiotemporally-homogeneous approach that was developed to account for observed spatial irregularities in ionospheric modeling \citep{walter2001, sparks2011b}. Uncertainty inflation using $R_\mathrm{irreg}$ is predicated upon on the unlikelihood of the $\chi^2$ goodness-of-fit statistic for a GPR fit. The  $\chi^2$ statistic, determined by the squared sum of whitened observations
\begin{equation}
\label{eq: chi2_def}
\chi^2 = y^T\Gamma^T\Gamma y,
\end{equation}
will follow a $\chi^2(d)$-distribution for covariances accurately describing GPs with $d$ degrees of freedom. To inflate GPR posterior uncertainties using $R_\mathrm{irreg}$, the posterior covariance provided in Equation \eqref{eq: gpr_cov}, but not the corresponding prior $K$ and $R$ matrices in the equation, is multiplied by the scalar value $R_\mathrm{irreg}^2$ given by
\begin{equation}
\label{eq: r_irreg_def}
R_\mathrm{irreg}^2 = \max \left\{ \frac{\chi^2} {\chi^2_{d, \mathrm{lower bound}}}, 1\right\}.
\end{equation}
Here, the value of $\chi^2_{d, \mathrm{lower bound}}$ is defined by the square of the following minimizer:
\begin{equation}
\label{eq: chi2_lower_bound_def}
\chi_{d, \mathrm{lower bound}} = \min\left\{\chi \in \mathbb{R}^+ \middle|
P \left( \left| X \right| \geq \frac
{\zeta \sqrt{Y}}
{\chi_{d, \mathrm{lower bound}}} 
\right) \leq P_\mathrm{threshold}
\right\}
\end{equation}
where $X$ and $Y$ are random variables with respective distributions that are standard unit normal and standard $\chi^2$ with $d$ degrees of freedom, and $\zeta$ and $P_\mathrm{threshold}$ are scalar values. Thus, multiplying all variances in Equation \eqref{eq:C_equal_K_plus_R} by $R_\mathrm{irreg}^2$ will achieve the ``worst'' case $\chi^2\left(d\right)$ scaled by $\zeta$ such that a probability of an unsampled threshold breach with probability threshold $P_\mathrm{threshold}$ can pass undetected. The parameters $\zeta$ and $P_\mathrm{threshold}$ have been set in ionospheric modeling applications to 5.592 and $10^{-10}$, respectively \citep{sparks2011b}, and will be assumed to be the same for all appropriate applications going forward. Inflating GPR-estimated variances by multiplying by $R_\mathrm{irreg}^2$ is viewed as a form of homogeneous inflation since all variances are inflated uniformly irrespective of any spatiotemporal components.


\section{Nonstationary Model Formulation}
\label{sec: models}

\begin{figure}[!htb]
\centering
\includegraphics[width=3.25in]{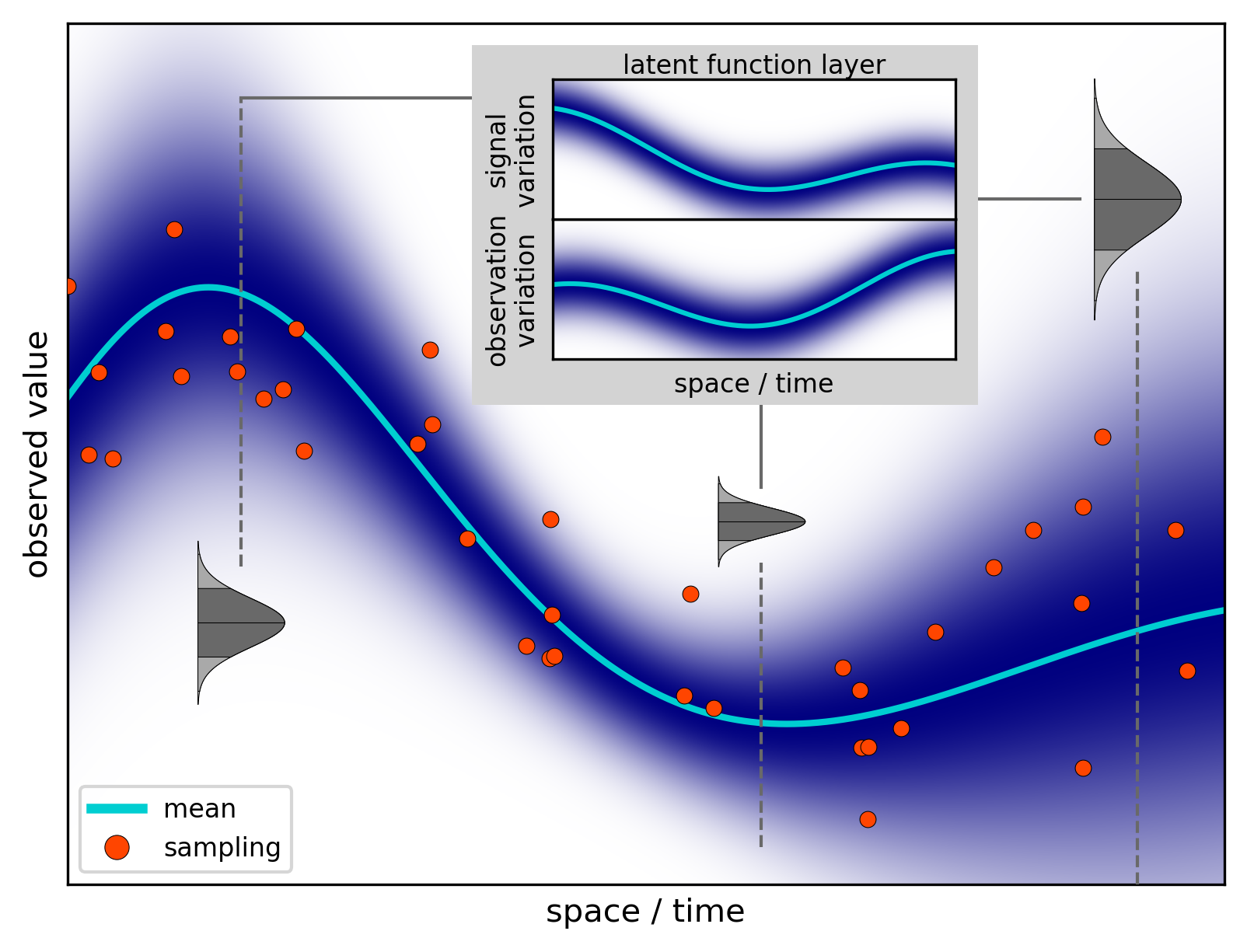}
\caption{Illustration of a nonstationary Gaussian process described by latent functions for signal and observation variation.}
\label{fig: layered_gaussian_process}
\end{figure}

Nonstationary GP modeling developed in this paper takes a novel approach that is designed with an integrity-based focus for safety-critical applications. Instead of assuming a variable length scale, nonstationarity is introduced into kernels through latent functions for signal and observation variation, schematically depicted in Figure \ref{fig: layered_gaussian_process}, that are discovered through a boosting procedure based on their relative errors, which are estimated using an extra layer of GP modeling in the partially-whitened domain. Methods for posterior uncertainty inflation are proposed and incorporated into an algorithm for nonstationary covariance boosting that conservatively inflates priors and posterior uncertainties to reduce the risks of both overfitting and overconfidence.

\subsection{Prior Covariance Boosting}
\label{subsec: boosting}

Suppose that the true prior covariance $C$ for a GP exists and is unknown. However, we will not assume that all of its components are unknown but that we have prior knowledge of its correlation structure.

\noindent
{\bf Assumption 1.} {\it Assume that the true correlation matrix, $\uprho$, is known and well-specified.
\hfill\BlackBox}
\medskip

Now, consider the following prior covariance
\begin{equation}
\label{eq: Cj_decompose}
C^{(j)} = \Sigma^{(j)} \uprho  \Sigma^{(j)} + R^{(j)}
\end{equation}
as an SPD approximation for $C$ with superscript denoting an index in a potential sequence of approximations, $\{C^{(j)}\}_{j=0}^J$. Here, the matrices $\Sigma^{(j)} = \mathrm{diag}\left(\sigma_\mathrm{signal}^{(j)} \right)$ and $R^{(j)} = \mathrm{diag}\left(\sigma_\mathrm{obs}^{(j)} \right)^2$ represent approximate signal variation and approximate observation variance, respectively, for $C^{(j)}$. Note that the approximate covariance in Equation \eqref{eq: Cj_decompose} assumes a true correlation matrix $\uprho$ that is shared by $C$, which is consistent with Assumption 1.

Let $\eta_\mathrm{signal}^{(j)}$ and $\eta_\mathrm{obs}^{(j)}$ represent the relative error of the $j$th approximation for signal and observation variation, respectively, which can be expressed in terms of the true variation as
\begin{equation}
\label{eq: eta_X_def}
\eta_\mathrm{X}^{(j)} = \frac{\sigma_\mathrm{X}}{\sigma_\mathrm{X}^{(j)}} - 1
\end{equation}
where ``X'' can represent the ``signal'' or ``obs'' subscripts. If the subsequent variation approximations accounted for the relative errors with a scalar learning rate of $\xi^{(j)} > 0$, then the $(j+1)$th approximations for signal and observation variation take the following form:
\begin{equation}
\label{eq: subsequent_sigma_X}
\sigma_\mathrm{X}^{(j+1)} = \left(1 + \xi^{(j)} \eta_\mathrm{X}^{(j)} \right) \sigma_\mathrm{X}^{(j)}.
\end{equation}
These variations can be expressed recursively in terms of approximation errors, learning rates, and initial weak learners, $\sigma_\mathrm{signal}^{(0)}$ and $\sigma_\mathrm{obs}^{(0)}$, as
\begin{equation}
\label{eq: subsequent_sigma_X}
\sigma_\mathrm{X}^{(j+1)} = \sigma_\mathrm{X}^{(0)} \prod_{i=0}^{j} \left(1 + \xi^{(i)} \eta_\mathrm{X}^{(i)} \right).
\end{equation}

When both $\eta_\mathrm{signal}^{(j)}$ and $\eta_\mathrm{obs}^{(j)}$ are negligible in magnitude over their entire domains, and $\xi^{(j)}$ is bounded above by 1, then $\sigma_\mathrm{signal}^{(j+1)}$ and $\sigma_\mathrm{obs}^{(j+1)}$ become suitable approximations for the true variations. As a result under these conditions, $C^{(j+1)}$ would be a suitable approximation, boosted from $C^{(0)}$, for the true prior covariance $C$.

It is worth noting that when any one of the relative errors $\eta^{(i)}$ is nonconstant for any $i$ for either signal or observation variation, then the resultant approximate prior covariance $C^{(j+1)}$ would be inhomogeneous and the resultant Gaussian process model would be nonstationary. On the other hand, if all relative errors, $\eta^{(i)}$, are constant and initial weak approximations are homogeneous, then the resultant Gaussian process would remain stationary.


\subsection{Partial Whitening Approximations}
\label{subsec: partial_whitening}

Let us consider an approximate prior covariance, $C^{(j)}$ as a whitener for observations, as specified in Equations \eqref{eq:whitening_definition} and \eqref{eq:mahalanobis}. Consider the random variable
\begin{equation}
\label{eq: y_hat_def}
\hat{y}^{(j)} := \Gamma^{(j)} y = \left(C^{(j)}\right)^{-1/2}y
\end{equation}
as a ``partial whitening'' of the observations $y$. We consider this whitening ``partial'' in this context under the impression that $C^{(j)}$ does not fully account for all variation described by $C$. Given $y \sim \mathcal{N}\left(0,C\right)$, The partially-whitened observations are also a GP with $\hat{y}^{(j)} \sim \mathcal{N}\left(0, \hat{C}^{(j)}\right)$
where
\begin{equation}
\label{eq: C_hat_def}
\hat{C}^{(j)} = \left(C^{(j)}\right)^{-1/2}C\left(C^{(j)}\right)^{-1/2}.
\end{equation}
Expressing the covariance for the partially-whitened observations in terms of the approximate prior covariance error matrix, $\Delta C^{(j)} = C - C^{(j)}$, yields
\begin{equation}
\label{eq: C_hat_with_DeltaC}
\hat{C}^{(j)} = \left(C^{(j)}\right)^{-1/2}\left(C^{(j)} + \Delta C^{(j)}\right)\left(C^{(j)}\right)^{-1/2}.
\end{equation}
Isolating $\Delta C^{(j)}$, we obtain
\begin{equation}
\label{eq: DeltaCj_form}
\Delta C^{(j)} = \left(C^{(j)}\right)^{1/2}\left(\hat{C}^{(j)} - I\right)\left(C^{(j)}\right)^{1/2}.
\end{equation}

Next, assumptions are made regarding the sign of prior covariances appearing in Equation \eqref{eq: DeltaCj_form}.

\medskip
\noindent
{\bf Assumption 2.} {\it Assume that the kernel function is non-negative, $k\geq 0$. We note that the commonly chosen Mat\'ern, exponential, or Gaussian radial basis function (RBF) kernel functions are all always positive, though negative kernels, such as the difference of Gaussians, do exist.
\hfill\BlackBox}
\medskip

\noindent
{\bf Assumption 3.} {\it All elements of the matrix $\mathrm{cov}\left[ \hat{y}^{(j)} \right]$ are non-negative.
\hfill\BlackBox}
\medskip

Assuming all covariances of partially-whitened observations are non-negative suggests no spatiotemporal anticorrelation. However, such lack of anticorrelation may not be the case when $C^{(j)}$ is a poor approximation for $C$, so Assumption 3 asserts that $C^{(j)}$ is sufficiently accurate enough of an approximation to avoid such a scenario.

Applying Assumptions 2 and 3 to the diagonal of Equation \eqref{eq: DeltaCj_form} enables the following approximate inequality from diagonal terms:
\begin{equation}
\label{eq: diag_deltaCj}
\left(\Delta C^{(j)}\right)_{i,i} \gtrsim C^{(j)}_{i,i} \left(\mathrm{var}\left[ \hat{y}^{(j)} \right] - 1 \right)_i.
\end{equation}
Expanding $\Sigma\uprho\Sigma - \Sigma^{(j)} \uprho\Sigma^{(j)}$ as a component of $\Delta C^{(j)}$ along the diagonal yields
\begin{equation}
\label{eq: delta_approx_1}
(\Delta\sigma_{\mathrm{signal},i}^{(j)})^2 + 2 \sigma_{\mathrm{signal}, i}^{(j)} \Delta \sigma_{\mathrm{signal}, i}^{(j)} + \Delta \left(\sigma^2_\mathrm{obs}\right)_i^{(j)}
\gtrsim
\left[ \left(\sigma_{\mathrm{signal}, i}^{(j)}\right)^2 + \left(\sigma^2_\mathrm{obs}\right)_i^{(j)} \right] \left(\mathrm{var}\left[ \hat{y}^{(j)} \right] - 1 \right)
\end{equation}
where $\Delta\sigma_\mathrm{signal}^{(j)} = \sigma_\mathrm{signal} - \sigma_\mathrm{signal}^{(j)}$ and $\Delta \left(\sigma^2_\mathrm{obs}\right)^{(j)} = \sigma^2_\mathrm{obs} - \left(\sigma^2_\mathrm{obs}\right)^{(j)}$ are the approximation errors for signal variation and the square of observation variation, respectively.

Approximate equality cases in Equations \eqref{eq: diag_deltaCj} and \eqref{eq: delta_approx_1} correspond to scenarios with $\mathrm{cov}\left[ \hat{y}^{(j)} \right]$ matrices with diagonal entries that are significantly larger than off-diagonal terms, or ones that are diagonally dominant. Such covariance matrices of partially-whitened observations would carry cross-correlations that are non-negative but nearly zero that suggest near spatiotemporal independence, which is a property of white noise. So, nearly accurate covariance approximations, $C^{(j)}$, likely correspond more to the approximate (``$\approx$'') cases of Equations \eqref{eq: diag_deltaCj} and \eqref{eq: delta_approx_1} and not nearly as much the strict inequality cases (``$>$'').


\subsection{Prior Covariance Relative Error Estimation}
\label{subsec: covariance_error_estimation}

Methods to estimate components of the prior covariance approximation errors,  $\Delta\sigma_\mathrm{signal}^{(j)}$ and $\Delta \left(\sigma^2_\mathrm{obs}\right)^{(j)}$, given the approximations $\sigma_\mathrm{signal}^{(j)}$ and $ \left(\sigma^2_\mathrm{obs}\right)^{(j)}$ for signal and observation noise, respectively, will rely on two primary avenues: (A) GP models for the variation of partially-whitened observations in multiple domains and (B) heuristic modeling. Several possible modeling approaches can be adopted to progress forward from approximations made in Section \ref{subsec: partial_whitening}; however, the approach outlined takes a mathematically na\"ive approach to the critical task of attributing variation in partially-whitened observations to signal and observation noise components of the prior covariance approximation error. As a result, the modeling presented here is best-suited for applications in which under-approximations for signal and observation variance are coincidental, meaning that irregularities in the input domain are assumed to enhance both types of variation.

First, we make a key heuristic assumption that enables the modeling of the observation component of the prior covariance approximation error.

\medskip
\noindent
{\bf Assumption 4.} {\it The observation variance approximation error can be well-approximated by a spatiotemporally inhomogeneous non-negative multiple of the magnitude in which the variance of the partially-whitened observations deviate from unity and of the approximate observation variance itself, as expressed by
\begin{equation}
\label{eq: delta_obs_model}
\Delta \left(\sigma^2_\mathrm{obs}\right)_i^{(j)} \approx \eta_{R,i}^{(j)} \left(\sigma^2_\mathrm{obs}\right)_i^{(j)}
\end{equation}
\begin{equation}
\label{eq: etaR_def}
\eta_{R,i}^{(j)} = \left(1 - a_i^{(j)} \right) \left( \mathrm{var}\left[ \hat{y}_i^{(j)} \right] - 1 \right)
\end{equation}
for some $0 \leq a_i^{(j)} \leq 1$ for all $i$.
}\hfill\BlackBox
\medskip

Here, $a^{(j)}$ is an unspecified latent function that represents the portion of the variance of the partially-whitened observations that deviate from unity, given by $\mathrm{var}\left[ \hat{y}^{(j)} \right] - 1$. The relative error for observation variation, $\eta_\mathrm{obs}$, referenced in the boosting model in Section \ref{subsec: boosting} is associated with the relative error for the $R$-matrix with observation variance along the diagonal, captured by $\eta_R$, as follows:
\begin{equation}
\label{eq: etaR_to_obs}
\eta_\mathrm{obs} = \sqrt{\eta_R + 1} - 1.
\end{equation}

Incorporating the observation variance approximation error in Equations \eqref{eq: delta_obs_model} and \eqref{eq: etaR_def} into Equation \eqref{eq: delta_approx_1} yields the following approximate inequality for signal variation approximation error:
\begin{equation}
\label{eq: delta_approx_2}
(\Delta\sigma_{\mathrm{signal},i}^{(j)})^2 + 2 \sigma_{\mathrm{signal}, i}^{(j)} \Delta \sigma_{\mathrm{signal}, i}^{(j)}
\gtrsim b_i^{(j)}
\end{equation}
where
\begin{equation}
\label{eq:bj_def}
b_i^{(j)} = \left[ \left(\sigma_{\mathrm{signal}, i}^{(j)}\right)^2 + a_i^{(j)} \left(\sigma^2_\mathrm{obs}\right)_i^{(j)} \right] \left(\mathrm{var}\left[ \hat{y}^{(j)} \right] - 1 \right)_i.
\end{equation}
The quadratic form of the approximate inequality in Equation \eqref{eq: delta_approx_2} is, then, solved by
\begin{equation}
\label{eq: delta_sigmaprocess_solution}
\Delta \sigma_{\mathrm{signal}, i}^{(j)} \gtrsim \eta_{\mathrm{signal},i}^{(j)} \sigma_{\mathrm{signal}, i}^{(j)}
\end{equation}
\begin{equation}
\label{eq:eta_process_definition}
\eta_{\mathrm{signal},i}^{(j)} = \sqrt{1 + \frac{b_i^{(j)}} {\left( \sigma_{\mathrm{signal},i}^{(j)}\right)^2} } - 1.
\end{equation}
When partial whitening accurately whitens the observations, meaning $C^{(j)}$ is an accurate covariance estimate, the variance of $\hat{y}_i^{(j)}$ will be nearly 1, which will result in negligible observation variance error, $\Delta\sigma_\mathrm{obs}^{(j)}\approx 0$. Based off of the rationale for strict approximations at the end of Section \ref{subsec: partial_whitening} from approximate inequalities, this case of accurate partial-whitening would also suggest to $\Delta\sigma_\mathrm{signal}^{(j)} \approx 0$. The quantity $1 - a^{(j)}$ represents the portion of the variance of the partially-whitened observations that deviate from unity to be attributed to observation variance approximation error.

To estimate $\Delta\sigma_\mathrm{obs}^{(j)}$ and $\Delta\sigma_\mathrm{signal}^{(j)}$ using Equations \eqref{eq: delta_obs_model} and \eqref{eq: delta_sigmaprocess_solution}, respectively, approximations for $\mathrm{var}\left[ \hat{y}^{(j)} \right]$ and $a^{(j)}$ are required. Motivated by these approximations, partially-whitened observations are modeled using two different GPs. A first approach for these GPs models would consider the variation of $\left(\hat{y}^{(j)}\right) ^ 2$ that is representative of $\mathrm{var}\left[ \hat{y}^{(j)} \right]$ but follows a $\chi^2(1)$ distribution when partial-whitening accurately whitens the observations. Naturally, consider the square of the partially-whitened observations mapped to the Gaussian domain, as expressed by the quantity
\begin{equation}
\label{eq: H_def}
Z_{\chi^2}^{(j)} = H\left(\left(\hat{y}^{(j)}\right) ^ 2 \right)
\end{equation}
with
\begin{equation}
\label{eq:H_def}
H( v ) = F_{\mathcal{N}(0,1)}^{-1}\left( F_{\chi^2(1)}(v) \right)
\end{equation}
where $F_D$ is the cumulative distribution function of the distribution $D$. Here, the function $H$ maps a $\chi^2(1)$ distribution to a unit normal distribution, $\mathcal{N}(0,1)$. If we let $\tilde{\uprho}$ be some SPD correlation matrix, then consider:
\begin{itemize}
\item $\tilde{Z}_{\chi^2}^{(j)}$ as a GP model for the quantity $Z_{\chi^2}^{(j)}$ using the prior covariance $\tilde{\uprho} + I$ to estimate $\mathrm{var}\left[ \hat{y}^{(j)} \right]$ in the Gaussian domain
\item $\tilde{Z}_{\mathcal{N}\left(0,1\right)}^{(j)}$ as a GP model for the partially-whitened observations $\hat{y}^{(j)}$ using the prior covariance $\tilde{\uprho} + I$
\end{itemize}
We note that $\hat{y}^{(j)}$ may not be Gaussian but will be nearly Gaussian when the partial-whitening is nearly accurate, suggesting the appropriateness for the GP model $\tilde{Z}_{\mathcal{N}\left(0,1\right)}^{(j)}$. Both $\tilde{Z}_{\chi^2}^{(j)}$ and $\tilde{Z}_{\chi^2}^{(j)}$ assume unitary signal noise, like white noise, but also na\"{\i}vely assume signal noise of the same magnitude.

Profiles of the GP $\tilde{Z}_{\chi^2}^{(j)}$, then, provide the primary foundation to approximate the variance of the partially-whitened observations that is central to relative errors of approximate prior covariance components in Equations \eqref{eq: delta_obs_model} and \eqref{eq: delta_sigmaprocess_solution}.

\medskip
\noindent
{\bf Assumption 5.} {\it
The approximation $\tilde{v}^{(j)}_\kappa \approx \mathrm{var}\left[ \hat{y}^{(j)} \right]$ is valid with
\begin{equation}
\label{eq: vj_approx}
\tilde{v}_\kappa^{(j)} = \frac{H^{-1}\left( \tilde{\omega}_{\chi^2, \kappa}^{(j)}  \right)}{H^{-1}\left(0\right)}
\end{equation}
\begin{equation}
\label{eq: omega_chi2_def}
\tilde{\omega}_{\chi^2, \kappa}^{(j)} = \mathbb{E}\left[ \tilde{Z}_{\chi^2}^{(j)}\right] + \kappa \sqrt{\mathrm{var}\left[ \tilde{Z}_{\chi^2}^{(j)}\right]}
\end{equation}
for some scalar parameter $\kappa \geq 0$. 
}\hfill\BlackBox
\medskip

At this juncture, our attention turns to potential instances in which $\mathrm{var}\left[ \hat{y}^{(j)} \right] < 1$. In such scenarios, the variance of the partially-whitened observation exceeds that of unitary white noise, which would be expected for observations whitened using the true prior covariance, $\mathrm{var}\left[ \hat{y} \right] = 1$, suggesting that at least one of the approximations $\sigma_\mathrm{signal}^{(j)}$ and $\sigma_\mathrm{obs}^{(j)}$ exceed their true counterparts $\sigma_\mathrm{signal}$ and $\sigma_\mathrm{obs}$, respectively. However, if these approximations yield an estimated variance of the partially-whitened observations with $\tilde{v}_\kappa^{(j)} < 1$, then it may be undesirable for a practitioner seeking to meet integrity standards to appropriately tune the approximate covariance by the decreasing the values of its components $\sigma_\mathrm{signal}^{(j)}$ and $\sigma_\mathrm{obs}^{(j)}$ at any spatiotemporal location knowing that it could result in deflated posterior uncertainties and, as a result, model overconfidence. This approach would be suitable as long as $\sigma_\mathrm{signal}^{(j)}$ and $\sigma_\mathrm{obs}^{(j)}$ are assumed to not overfit measurements.  Thus, we will assume that the approximate covariance components $\sigma_\mathrm{signal}^{(j)}$ and $\sigma_\mathrm{obs}^{(j)}$ are chosen to be approximate lower bounds for their the true respective covariance components. In that regard, the approximate covariance $C^{(j)}$ can be considered a ``nominal'' covariance prior that has yet to be tuned, specifically by inflation, to account for any potential irregularities characterized by the variance of partially-whitened observations that exceeds unity.

\medskip
\noindent
{\bf Assumption 6.} {\it
Assume that $\sigma_\mathrm{signal}^{(j)} \lesssim \sigma_\mathrm{signal}$ and $\sigma_\mathrm{obs}^{(j)} \lesssim \sigma_\mathrm{obs}$ on the entire domain. Then, it follows that $\mathrm{var}\left[ \hat{y}^{(j)} \right] \gtrsim 1$ with appropriateness to require the approximation $\tilde{v}_\kappa^{(j)}$ at or above 1. As a result, the approximation $\tilde{V}^{(j)}_\kappa \approx \mathrm{var}\left[ \hat{y}^{(j)} \right]$ is valid with
\begin{equation}
\label{eq: Vj_approx}
\tilde{V}_\kappa^{(j)} = \Psi \left(\tilde{v}_\kappa^{(j)} ; 1, \beta \right)
\end{equation}
where
\begin{equation}
\label{eq: softplus_def}
 \Psi(u; \alpha, \beta) = \log \left[1 + \frac{1}{\beta}\exp\left(\beta \left(u - \alpha\right)\right)\right] + \alpha.
\end{equation}
}\hfill\BlackBox
\medskip

\begin{figure}[!htb]
\centering
\includegraphics[width=2.5in]{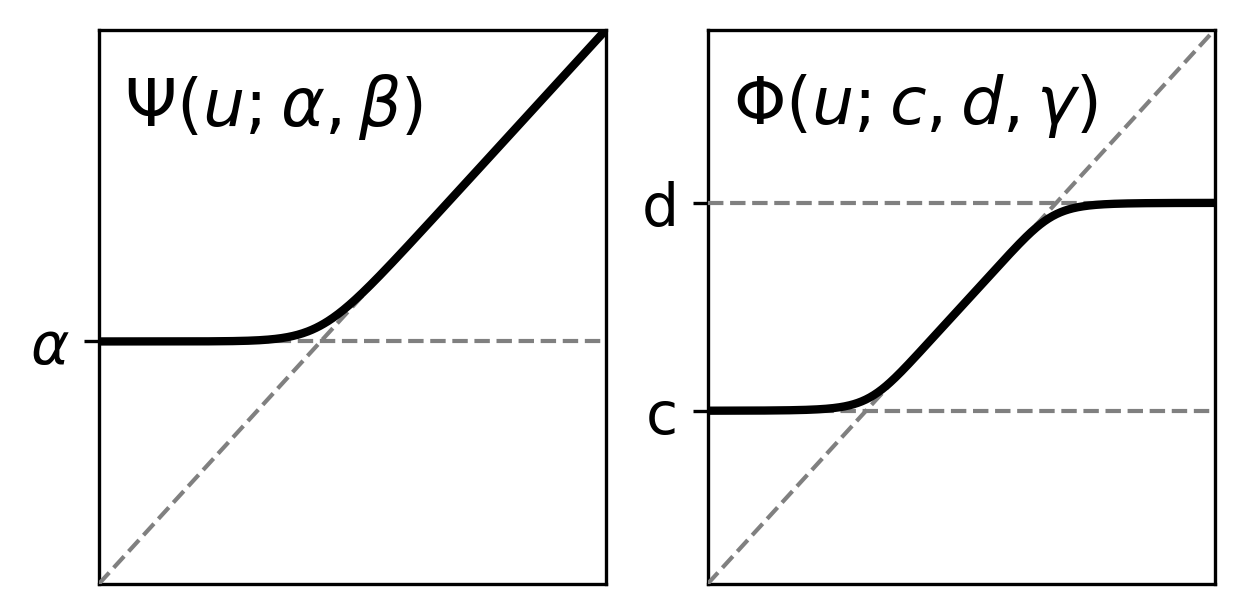}
\caption{Smooth thresholding functions $\Psi$ ({\it left}) and $\Phi$ ({\it right}) guided by threshold parameter values and the identity function.}
\label{fig:smoothing_functions}
\end{figure}

The scaled and shifted softplus function $\Psi$, also recognized as a smooth rectified linear unit function and represented in the top panel of Figure \ref{fig:smoothing_functions}, maintains the value of $\tilde{V}_\kappa^{(j)}$ above 1 in a smooth and continuously differentiable fashion while approximating the identity function for values significantly larger than 1. When $\kappa > 0$, $\tilde{\omega}_{\chi^2, \kappa}^{(j)}$ represents an upper bound on a confidence interval of the GP posteriors for $\tilde{Z}_{\chi^2}^{(j)}$ as prescribed by $\kappa$. The value of $\tilde{\omega}_{\chi^2, \kappa}^{(j)}$ simply reduces to the posterior expected value when $\kappa = 0$. Since it is in the Gaussian domain, applying $H^{-1}$ to model $\tilde{Z}_{\chi^2}^{(j)}$ values maps them to the $\chi^2(1)$ domain, the more appropriate space for variance estimation. 

Our final assumption provides means to estimate the last remaining component of relative errors of covariance approximations, $a^{(j)}$.

\medskip
\noindent
{\bf Assumption 7.} {\it
The quantity $\tilde{a}_\kappa^{(j)}$ given by
\begin{equation}
\label{eq: a_kappa_approx}
\tilde{a}_\kappa^{(j)} = 1 - \sqrt{\tilde{r}_\kappa^{(j)}}
\end{equation}
\begin{equation}
\label{eq: rj_ratio}
\tilde{r}_\kappa^{(j)} = \Phi \left(
\frac{\left( \tilde{\omega}_{\mathcal{N}\left(0,1\right), \kappa}^{(j)} \right)^2 + \left\Vert \zeta_k \nabla \left( \mathbb{E}\left[\tilde{Z}_{\mathcal{N}(0,1)}^{(j)}\right] \right) \right\Vert^2}
{\tilde{V}_\kappa^{(j)} H^{-1}\left(0\right)}; 
0, 1, \gamma \right)
\end{equation}
\begin{equation}
\label{eq: omega_n01_def}
\tilde{\omega}_{\mathcal{N}\left(0,1\right), \kappa}^{(j)} = \mathbb{E}\left[\tilde{Z}_{\mathcal{N}(0,1)}^{(j)}\right] + \kappa \,  \mathrm{sign}\left(  \mathbb{E}\left[\tilde{Z}_{\mathcal{N}(0,1)}^{(j)}\right] \right) \sqrt{\mathrm{var} \left[\tilde{Z}_{\mathcal{N}(0,1))}^{(j)}\right]}
\end{equation}
where $\zeta_k > 0$ is some correlation-specific scaling factor and
\begin{equation}
\label{eq: softthresholding_def}
 \Phi(u; c, d, \gamma) = \frac{1}{2} \left[ \frac{2u - c - d}
{\left( 1 + \left(\frac{2 u-c-d}{d-c} \right)^\gamma \right)^{1/\gamma}}
+ c + d \right]
\end{equation}
provides a reasonable heuristic estimate for $a^{(j)}$ when $\left|\eta_{R}^{(j)}\right| \gg 1$ and $\left| \eta_\mathrm{signal}^{(j)} \right| \gg 1$ are predominantly coincidental spatiotemporally with the same sign.
}\hfill\BlackBox
\medskip

The smooth function $\Phi$ thresholds input values between scalar parameters $c$ and $d$ with a steepening coefficient of $\gamma$, as shown in the bottom panel of Figure \ref{fig:smoothing_functions}. The quantity $\tilde{\omega}_{\mathcal{N}(0,1), \kappa}^{(j)}$ represents a $\kappa$-dependent confidence bound for $\tilde{Z}_{\mathcal{N}(0,1)}^{(j)}$ that depends on the sign of the expected value, which results in magnitudes above the model's expected value for $\kappa > 0$. The scaling factor, $\zeta_k$, for the exponential decay and squared exponential decay correlation functions are $\zeta_\mathrm{OU} = \sqrt{2\tilde{L}}$ and $\zeta_\mathrm{RBF} = \sqrt{2} \tilde{L}$, respectively, where $\tilde{L}$ is the length scale for the GPR fit for $\tilde{Z}_{\mathcal{N}(0,1)}^{(j)}$. These values scale gradient magnitudes to be comparable to the magnitude of $\tilde{Z}_{\mathcal{N}(0,1)}^{(j)}$.

The ratio $\tilde{r}_\kappa^{(j)}$ quantifies the portion of the variance of the partially-whitened observations that can be solely attributed to observation variance as opposed to signal variance. For example, a nearly zero-valued $\tilde{r}_\kappa^{(j)}$ arises when there is little-to-no variation of $\tilde{Z}_{\mathcal{N}(0,1)}^{(j)}$ relative to $\tilde{Z}_{\chi^2}^{(j)}$, suggesting that the variation of $\hat{y}^{(j)}$ is predominantly uncorrelated. On the other hand, values of $\tilde{r}_\kappa^{(j)}$ nearing 1 indicates predominantly highly-correlated signal-oriented variation of $\hat{y}^{(j)}$.

With a heuristic nature, the value of $\tilde{r}_\kappa^{(j)}$ may be somewhat inaccurate in some scenarios but still be ``reasonable,'' per Assumption 7, due to the expectation that larger relative errors of the covariance components $\eta_{R}^{(j)}$ and $\eta_\mathrm{signal}^{(j)}$ are co-located. In general, any mismodeling of $a^{(j)}$ may wrongly attribute signal variation approximation errors to observation variation approximation errors and vice versa; however, the spatiotemporal coincidence of both types of approximation errors may lessen the impact of such mismodeling in practice. For example, an iterative refinement of covariance approximations made with minor incremental changes from successive approximation errors could make incorrect attributions between signal and observation variation approximation errors at a single iteration but still ultimately converge towards a fairly accurate approximation of the true prior covariance.

\subsection{Post-Fit Uncertainty Inflation via Effective Standard Error}
\label{subsec: postfit_inflation}

After the prior covariance matrix is boosted using relative error estimates, we develop an approach for posterior uncertainty inflation as an inhomogeneous alternative to $R_\mathrm{irreg}$ inflation to meet integrity standards by inflating: (A) the tuned prior covariance for further inflation post-fit and (B) the posterior covariance predicted at $x^*$. This post-fit inflation leverages our models for latent signal and observation variation errors discussed in Sections \ref{subsec: boosting}$-$\ref{subsec: covariance_error_estimation} and predominantly lies in the well-applied formulations for the standard errors of the mean and variance, which depend on sample numbers. For GPR, the sample number used for a fit is not translatable to the independent sampling of a distribution due to expected non-zero correlations across the samples. Instead, we will assume that an effective sample number, $N_\mathrm{eff}$, for a GPR fit can be approximated at each spatial location which would be analogous to the confidence specified by that independent sample number from a Gaussian distribution, as captured by the concept of effective sample size associated with design effect \citep{kish1965, thiebaux1984}.

The essence of inhomogeneous post-fit inflation for nonstationary GPR methods outlined in this study can be thought of as by applying an ``effective standard error'' for both the mean and the variance to inflate post-fit uncertainties using an effective sample number in different contexts. First, the $\tilde{\omega}$-values prescribed in Equations \eqref{eq: omega_chi2_def} and \eqref{eq: omega_n01_def} are $\kappa$-specified slices of GP posteriors in whitened domains. With $N_\mathrm{eff}$ representing an approximation for the effective sample number for a GPR fit at $x^*$, then choosing the $\kappa$ values to be
\begin{equation}
\label{eq: kappa_Neff}
\kappa_\mathrm{infl} = \frac{z_\mathrm{infl}}{\max\left\{\sqrt{N_\mathrm{eff}}, \epsilon_\mathrm{eff}\right\}}
\end{equation}
formulates $\tilde{\omega}$-values such that their means are offset by multiples of the effective standard error for the mean for some scalar parameters $z_\mathrm{infl}, \epsilon_\mathrm{eff} > 0$. Here, the value $\epsilon_\mathrm{eff}^2$ serves as a floor value for $N_\mathrm{eff}$ so that $\kappa_\mathrm{infl}$ does not diverge to $\infty$ in heavily undersampled areas.  Next, scaling $\kappa_\mathrm{infl}$ by $\sqrt{2}$ yields a $z_\mathrm{infl}$-multiple of the effective standard error coefficient for the variance, which appropriately inflates the GP posterior variance in the form in Equation \eqref{eq: gpr_cov} to yield
\begin{equation}
\label{eq: gpr_var_infl}
\sigma_\mathrm{infl}^2 = \left(1 + \sqrt{2}\kappa_\mathrm{infl} \right) \mathrm{diag}\left(\mathrm{cov}\left[ y^* \right]\right).
\end{equation}
So, the spatiotemporally-inhomogneous $\kappa_\mathrm{infl}$ is used twice for post-fit inflation: once for sampling the GPR fits to partially-whitened observations further out from the expected value and a second time to inflate the posterior variance in a multiplicative fashion.

Though design effect models often account for correlations when estimating effective sample numbers, this study takes a more simplified approach that only factors in spatial proximity. For all nonstationary GP modeling presented here, $N_\mathrm{eff}$ will be estimated using a scaled Gaussian convolution with a representation of a sum of Dirac delta functions at the spatiotemporal location of each sampled measurement that takes the form
\begin{equation}
\label{eq: N_eff_specific}
N_\mathrm{eff}(u) = \sum_{i=1}^N \mathrm{exp}\left(\frac{-\lVert u - x_i \rVert^2} {L_\mathrm{eff}^2} \right)
\end{equation}
where $L_\mathrm{eff}$ represents the Gaussian length scale scaled by $\sqrt{2}$ and $u$ represents an arbitrary spatiotemporal state. It is important to note that, through this coarse design, $N_\mathrm{eff}$ is invariant to the correlation function, $\rho$, and determined without the clustering of observations accounted for.

The specifics of how the approximate prior covariances are proposed to be used distinctly from these post-fit inflated uncertainties in practice is outlined in the following section.


\subsection{Nonstationary Covariance Boosting Algorithm}
\label{subsec: covariance_inflation_algorithm}

Practical incorporation of the prior covariance boosting methods and relative error approximations in Sections \ref{subsec: boosting}$-$\ref{subsec: postfit_inflation} into GPR models requires specific attention to both potential overfitting and model overconfidence. We present an approach that relies on four different prior covariance approximations:
\begin{enumerate}[noitemsep]
\item An initial approximate prior covariance, $C^{(0)}$, that is assumed to be comprised of an accurate correlation matrix, $\uprho$, and signal and observation variation approximations that do not overapproximate true noise, given by $\sigma_\mathrm{signal}^{(0)} \lesssim \sigma_\mathrm{signal}$ and $\sigma_\mathrm{obs}^{(0)} \lesssim \sigma_\mathrm{obs}$, respectively.
\item The prior covariance that best whitens the observations as determined a by boosting process guided by analysis in Section \ref{subsec: boosting}$-$\ref{subsec: covariance_error_estimation}, denoted by $C^{(j)}$.
\item A prior covariance used for GPR fitting, denoted by $C_\mathrm{fit}$, that applies a smooth thresholding to the latent variation functions from $C^{(j)}$ as means to prevent overfitting.
\item A prior covariance used post-fitting to estimate inflated GP posterior uncertainties to prevent overconfident posteriors using methods outlined in Section \ref{subsec: postfit_inflation}, denoted by $C_\mathrm{infl}$.
\end{enumerate}

\begin{figure}[!htb]
\centering
\includegraphics[width=5in]{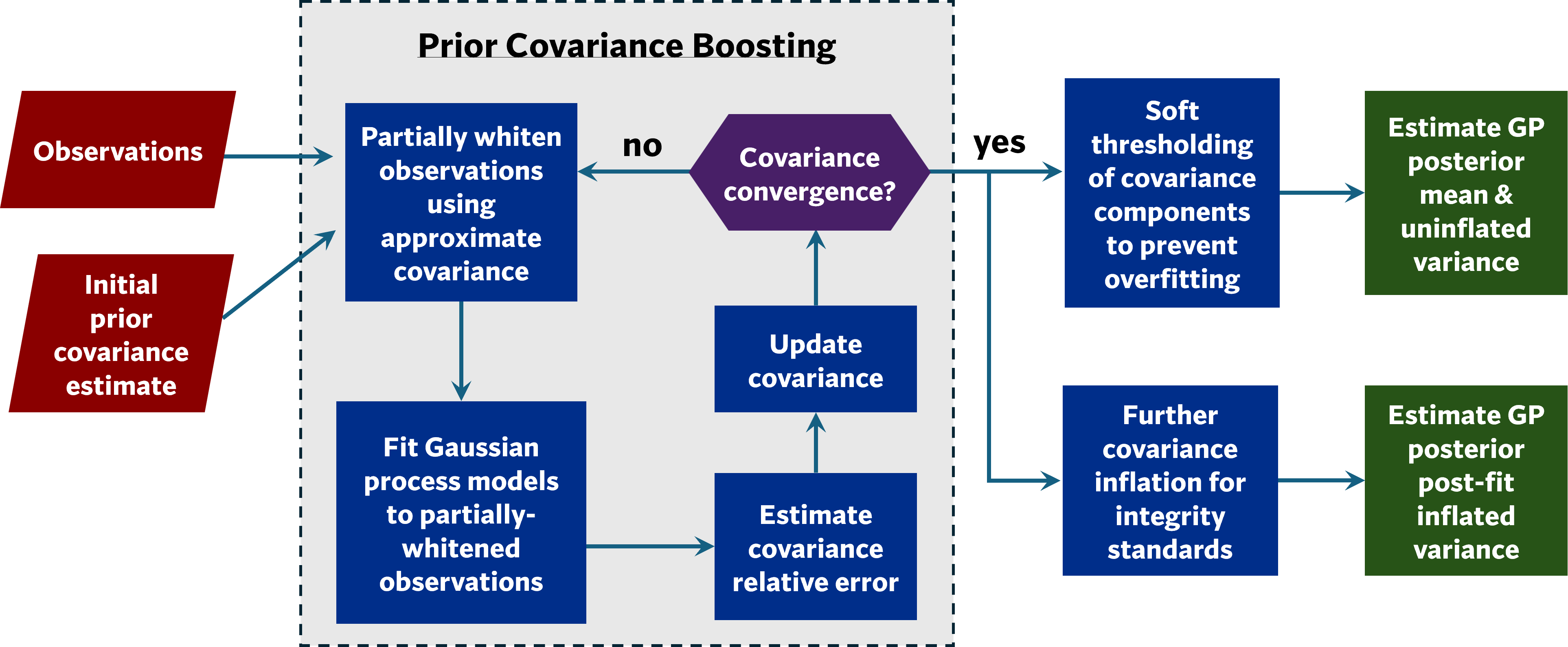}
\caption{Schematic diagram of high-level procedures for Covariance-Boosted Gaussian Process (CBGP) methods outlined in Algorithm \ref{alg: cov_infl_alg}.}
\label{fig:schem_diagram}
\end{figure}

Algorithm \ref{alg: cov_infl_alg} outlines a methodology to estimate GP posterior means ($\mu$), uncertainties ($\sigma$), and post-fit-inflated uncertainties ($\sigma_\mathrm{infl}$) using these four model prior covariances, which we will further refer to as a Covariance-Boosted Gaussian Process (CBGP). Figure \ref{fig:schem_diagram} illustrates the components of Algorithm \ref{alg: cov_infl_alg} as a process diagram. 

Algorithm \ref{alg: cov_infl_alg} is concisely structured so that its \texttt{while} loop computes all three of the prior covariances used for the computation of GP posteriors, $C^{(j)}$, $C_\mathrm{fit}$, and $C_\mathrm{infl}$. First, $C^{(j)}$ is computed through a boosting procedure occurring under the conditions of when \texttt{IsConverged} is set to \texttt{False}. Boosting iterations are designed in a way similar to a gradient descent method where incremental changes to $\sigma_\mathrm{signal}$ and $\sigma_\mathrm{obs}$ are made proportional to the relative errors, $\eta$, as scaled by a learning rate, $\xi$, that is not directly incorporated into convergence criteria of the iterative method, which occurs when the maximum relative error component, $\delta$, no longer exceeds a prescribed tolerance, $\varepsilon$. Several learning rate choices may be appropriate, including a constant one; however, in this study, we take a dynamic learning rate of
\begin{equation}
\label{eq: learning_rate}
\xi = \min\left\{\frac{\xi_0} {(1+\delta)^2}, 1\right\}
\end{equation}
that assumes small values for large relative errors and does not exceed a value of 1. It is worth noting that the iterative procedure in Algorithm \ref{alg: cov_infl_alg} specifies for the smooth thresholding of $\hat{y}^{(j)}$ in practice using the function $\Psi$, so its magnitude does not exceed a prescribed value of $z_{\mathrm{T}}$. During boosting iterations, GPR fits are performed to partially-whitened observations, but GP posteriors are not computed for observations in the input domain.

The second stage of Algorithm \ref{alg: cov_infl_alg} that follows the completion of prior covariance boosting, characterized by when \texttt{IsConverged} is \texttt{True} and \texttt{isFitted} is \texttt{False}, guides the GPR fitting of the observations. Posterior means, $\mu$, and uncertainties, $\sigma$, are computed after further refining $C^{(j)}$ to prevent potential overfitting by the smooth thresholding of $\sigma_\mathrm{signal}^{(j)}$ and $\sigma_\mathrm{obs}^{(j)}$. A signed version of the softplus function $\Psi$ bound the $\sigma$-components of the refine prior covariance $C_\mathrm{fit}$, denoted by $\sigma_\mathrm{signal}^{(j)}$ and $\sigma_\mathrm{obs}^{(j)}$, by above by the scalar parameters $\sigma_\mathrm{s, max}$ and $\sigma_\mathrm{o, max}$, respectively.

The third and final stage of Algorithm \ref{alg: cov_infl_alg} computes the inflated uncertainty $\sigma_\mathrm{infl}$ when both \texttt{IsConverged} and \texttt{isFitted} are \texttt{True}. Inhomogeneous inflation methods are based on ``effective standard errors'' that are outlined in Section \ref{subsec: postfit_inflation}. At this stage, further inflation of latent signal and observation variation quantified by $\sigma_\mathrm{signal, infl}$ and $\sigma_\mathrm{obs, infl}$, respectively, is performed using a single iteration with a learning rate of 1.

Since the SPD matrix $\uprho$ serves as their correlation component, all three of the prior covariances $C^{(j)}$, $C_\mathrm{fit}$, and $C_\mathrm{infl}$ that are used to provide CBGP posteriors are also SPD matrices, given that $C^{(0)}$ is SPD.

\begin{algorithm}
\caption{Nonsationary Prior Covariance Boosting}
\scriptsize
\label{alg: cov_infl_alg}
\begin{algorithmic}
\State \textbf{Given}: initial covariance components $\sigma_\mathrm{signal}^{(0)}$, $\sigma_\mathrm{obs}^{(0)}$, $\uprho$
\State \textbf{Parameters}: $\varepsilon$, $z_\mathrm{infl}$, $z_T$, $\sigma_\mathrm{s, max}$, $\sigma_\mathrm{o, max}$, $\gamma_\Psi$, $\gamma_\Phi$, $\kappa_0$, $\uprho_\eta$
\State $ \tilde{x} \gets [x^T, x^{*T}]^T $    \Comment{Concatenate training and (potentially) unsampled spatiotemporal states}
\State $j \gets 0$     \Comment{Initialize iteration count}
\State $\kappa \gets \kappa_0$ 
\State $\delta \gets 1 + \varepsilon$    \Comment{Initialize max relative error difference}
\State $N_\mathrm{eff} \gets \mathrm{ApproxEffectiveSampleNum}\left(\tilde{x},\dots\right)$   \Comment{Effective sample number estimate at $\tilde{x}$}
\State $\sigma_\mathrm{signal} \gets \sigma_\mathrm{signal}^{(0)} \vec{1} $  \Comment{Weak learner for signal variation}
\State $\sigma_\mathrm{obs} \gets \sigma_\mathrm{obs}^{(0)} \vec{1} $   \Comment{Weak learner for observation variation}
\State $\texttt{IsConverged} \gets \texttt{False}$   \Comment{Specifier for boosting convergence}
\State $\texttt{IsFitted} \gets \texttt{False}$     \Comment{Specifier for whether GPR fit has been performed}
\State $\texttt{doIteration} \gets \texttt{True}$   \Comment{Specifier to iterate}
\While{\texttt{doIteration}}
    \If{not \texttt{IsConverged}}
        \State $C^{(j)}  \gets \mathrm{ConstructCovariance}\left(\sigma_\mathrm{signal}, \sigma_\mathrm{obs}, \rho\right)$
        \State $\Gamma \gets \left(C^{(j)}\right)^{-1/2}$      \Comment{Whitening matrix}
        \State $\hat{y} \gets \Gamma y$       \Comment{Partial whitening of observations}
        \State $\hat{Y} \gets \Phi \left( \hat{y}; -z_T, z_T, \gamma_\Phi \right)$       \Comment{Smooth thresholding of whitened observations}
        \State $ \mu_{\chi^2}, \sigma_{\chi^2}  \gets \mathrm{GPRFit}\left(\tilde{x}, H( \hat{Y})^2, C = \uprho_\eta + I, K = \uprho_\eta, R = I \right)$
        \State $ \mu_{\mathcal{N}(0,1)}, \sigma_{\mathcal{N}(0,1)}, \nabla\left( \tilde{\mu}_{\mathcal{N}(0,1)} \right) \gets \mathrm{GPRFit}\left(\tilde{x}, \hat{Y}, C = \uprho_\eta + I, K = \uprho_\eta, R = I \right)$
    \EndIf
    \State $\omega_{\chi^2} \gets \mu_{\chi^2} + \kappa \sigma_{\chi^2} $
    \State $\omega_{\mathcal{N}(0,1)} \gets \mu_{\mathcal{N}(0,1)} + \kappa \sigma_{\mathcal{N}(0,1)} $
    \State $V \gets \Psi \left(H^{-1}(\omega_{\chi^2}) / H^{-1}(0); 1, \gamma_\Psi \right)$
    \State $r \gets $ Equation \eqref{eq: rj_ratio}
    \State $a \gets 1 - \sqrt{r}$
    \State $b \gets \left( \sigma_\mathrm{signal}^2 + a \sigma_\mathrm{obs}^2 \right) \left(V - 1\right) $
    \State $\eta_R \gets \left(1 - a\right) \left(V - 1 \right) $
    \State $\eta_\mathrm{obs} \gets \sqrt{\eta_R + 1} - 1 $
    \State $\eta_\mathrm{signal} \gets \sqrt{1+\frac{b}{\sigma_\mathrm{signal}^2}} - 1$
    \If{not \texttt{IsConverged}}
        \State $\delta \gets \mathrm{max}\left\{\eta_\mathrm{obs}, \eta_\mathrm{signal}\right\} $   \Comment{Maximum relative error difference}
        \State $ \xi \gets \mathrm{UpdateLearningRate}\left(\delta, \dots\right)$
        \State $\sigma_\mathrm{signal} \gets \sigma_\mathrm{signal} \left(1 + \xi \eta_\mathrm{signal} \right)$    \Comment{Update signal variation function}
        \State $\sigma_\mathrm{obs} \gets \sigma_\mathrm{obs} \left(1 + \xi \eta_\mathrm{obs} \right)$    \Comment{Update observation variation function}
        \If{$\delta < \varepsilon$}
            \State $\texttt{IsConverged} \gets \texttt{True}$
        \EndIf
    \EndIf
    \If{\texttt{IsConverged}}
        \If{not \texttt{IsFitted}}   \Comment{Compute posteriors}
            \State $\sigma_\mathrm{signal, fit} \gets -\Psi(-\sigma_\mathrm{signal}, -\sigma_\mathrm{s, max}, \gamma_\Psi)$   \Comment{Smooth thresholding of signal variation}
            \State $\sigma_\mathrm{obs, fit} \gets -\Psi(-\sigma_\mathrm{obs}, -\sigma_\mathrm{o, max}, \gamma_\Psi)$   \Comment{Smooth thresholding of observation variation}
            \State $C_\mathrm{fit}, K_\mathrm{fit}, R_\mathrm{fit}  \gets \mathrm{ConstructCovariance}\left(\sigma_\mathrm{signal, fit}, \sigma_\mathrm{obs, fit}, \rho\right)$
            \State $\mu, \sigma \gets \mathrm{GPRFit}\left( x^*, y, C_\mathrm{fit}, K_\mathrm{fit}, R_\mathrm{fit} \right)$
            \State $\texttt{IsFitted} \gets \texttt{True}$    \Comment{Designate completion of fitting}
            \State $\kappa \gets \frac{z_\mathrm{infl}} {\max\{N_\mathrm{eff}, \epsilon_\mathrm{eff}\}}$  \Comment{Determine $\kappa$ for $\sigma_\mathrm{infl}$ computation in following iteration}
        \Else     \Comment{$\sigma_\mathrm{infl}$ computation}
            \State $\sigma_\mathrm{signal, infl} \gets \sigma_\mathrm{signal} \left(1 + \eta_\mathrm{signal} \right)$    \Comment{Further inflation of signal variation}
            \State $\sigma_\mathrm{obs, infl} \gets \sigma_\mathrm{obs} \left(1 + \eta_\mathrm{obs} \right)$    \Comment{Further inflation of observation variation}
            \State $C_\mathrm{infl}, K_\mathrm{infl}, R_\mathrm{infl}  \gets \mathrm{ConstructCovariance}\left(\sigma_\mathrm{signal, infl}, \sigma_\mathrm{obs, infl}, \rho\right)$
            \State $\sigma_\mathrm{infl} \gets \mathrm{GPRFit}\left( x^*, y, C_\mathrm{infl}, K_\mathrm{infl}, R_\mathrm{infl} \right)$
            \State $\sigma_\mathrm{infl} \gets \sqrt{1 + \sqrt{2}\kappa} \sigma_\mathrm{infl}$
            \State $\texttt{doIteration} \gets \texttt{False}$   \Comment{Specify for the termination of the iterative scheme}
        \EndIf
    \EndIf
    \State $j \gets j+1$    \Comment{Update iteration counter}
\EndWhile
\State \textbf{return}: $\mu$, $\sigma$, $\sigma_\mathrm{infl}$
\end{algorithmic}
\end{algorithm}


\section{Computational Results}
\label{sec: results}

This section provides results for out-of-sample computational tests of the CBGP modeling framework profiled in Section \ref{sec: models} for both accuracy and integrity. The first application for the CBGP framework is a synthetic one-dimensional problem under ``nominal'' and ``disturbed'' conditions that showcases features of our models and their ability to account for spatiotemporal irregularities injected into a stationary system. Next, computational experiments are performed for two benchmark problems with comparisons to other stationary, heteroskedastic, and nonstationary GP models. Our final and primary application problem fits ionospheric measurements over South America collected during disturbed space weather conditions in both locally planar and regional contexts for four five-day space weather storms. 

All four of the computational test cases, including both synthetic and real scenarios, are noted to predominantly align with assumptions made in Section \ref{sec: models}. Notably, spatiotemporal irregularities present in the observations are largely characterized by concurrent heightened signal and observation noise. Furthermore, spatiotemporal correlation lengthscales are seemingly homogeneous enough to obtain correlation matrices, $\uprho$, that are considered sufficiently accurate for approximate prior covariances in the form of Equation \eqref{eq: Cj_decompose}.

Initial prior covariance approximations that serve as weak learners for CBGP modeling are assumed to be stationary and to describe ``nominal'' conditions for each testing problem that do not significantly over-approximate neither signal nor observation variation, which are then boosted using methods outlined in Algorithm \ref{alg: cov_infl_alg}. These initial prior covariances are expected to underfit and underestimate uninflated uncertainties for any irregularities sampled in the observations, which consistent with Assumption 6.

All out-of-sample testing will be subject to a ``three-nines'' integrity standard that is consistent with ionospheric error integrity bounding used by WAAS \citep{waas2025}, meaning that the GPR fit errors of 99.9\% of deprived observations will be within the 99.9\% confidence interval prescribed by the post-fit inflated GPR variance, or approximately 3.29 standard deviations from the expected value \citep{piedad2001}. Thus, three-nines integrity standards are met by GPR fits when at least 99.9\% of out-of-sample errors standardized by post-fit inflated posterior uncertainties do not exceed 3.29. These inflated uncertainties for existing GP models refer to $R_\mathrm{irreg}$ multiplied by the posterior variance given by the diagonal of Equation \eqref{eq: gpr_cov}. For CBGP methods outlined in this paper, these inflated uncertainties are simply $\sigma_\mathrm{infl}$.

Details and parameterizations for all computational tests are provided in Appendix \ref{app: params}.


\subsection{One-Dimensional Simulated Scenario}
\label{subsec: simulated_results}

The efficacy of our CBGP modeling framework is, first, demonstrated by a synthetic scenario characterized by spatiotemporal irregularities injected into a stationary GP process. This application problem is designed and parameterized to be a one-dimensional version of the detrended stationary GP model used by WAAS for ionospheric estimation, as described in \citet{sparks2011a}. A nominal stationary GP with an exponential decay kernel is defined on the unitless one-dimensional interval $\left[-20000, 20000\right]$, comparable to local neighborhoods for ionospheric modeling ranging from 800 km to 2,100 km, using a covariance length scale ($L_\rho = 8,000$), signal variation parameter ($\sigma_{\mathrm{signal},0} = \sqrt{0.91}$), and observation variation parameter ($\sigma_{\mathrm{obs},0} =0.3$) adopted from WAAS' model \citep{sparks2011a}. 

\begin{figure}[htb!]
 \centering
 \includegraphics[width=5in]{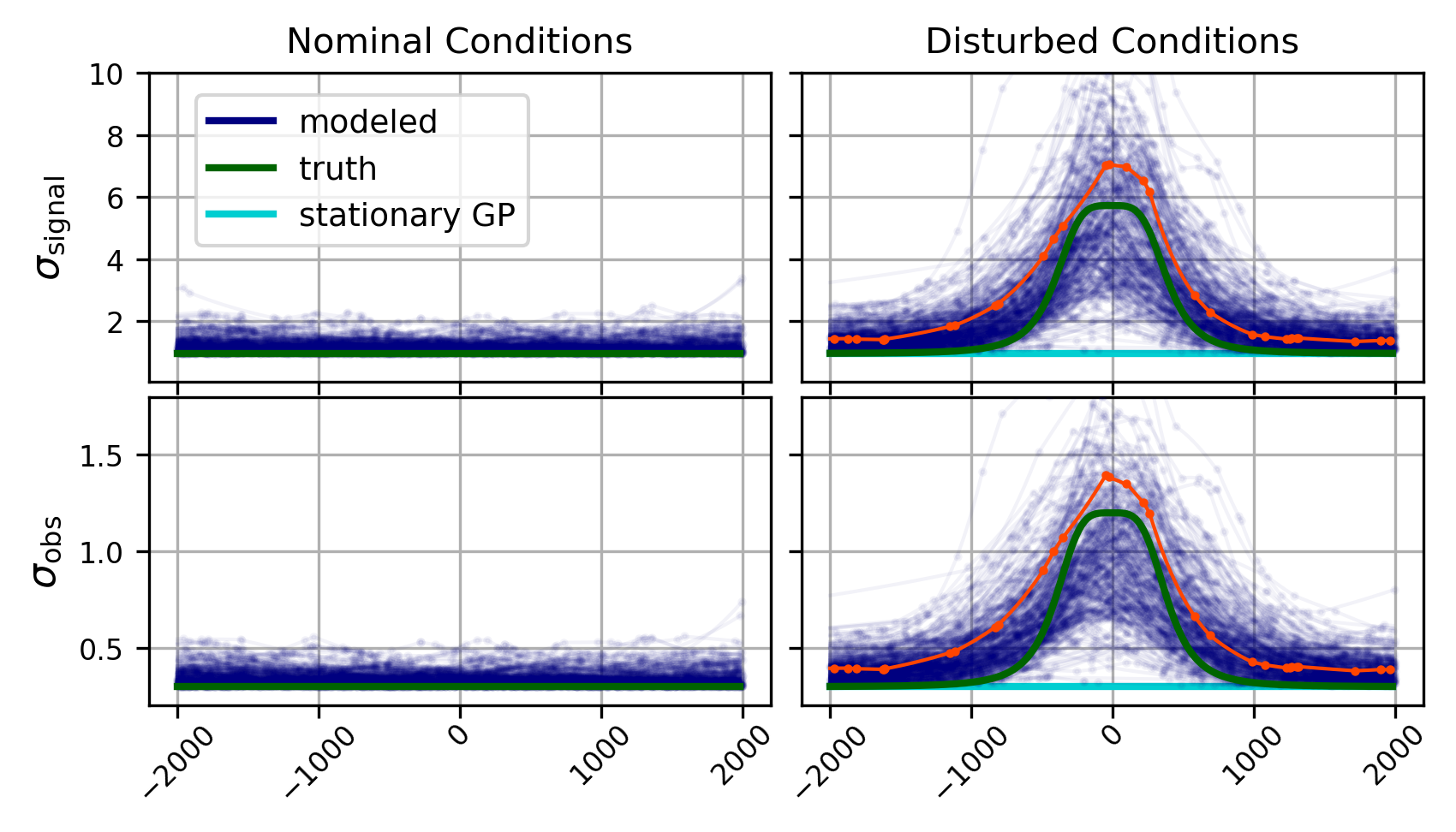}
\caption{CBGP-estimated latent $\sigma_\mathrm{signal}$ (\textsl{top}) and $\sigma_\mathrm{obs}$ (\textsl{bottom}) functions under nominal (\textsl{left}) and disturbed (\textsl{right}) conditions for the first 250 one-dimensional simulations along with respective true functions described by Equation \eqref{eq: synthetic_sigma} (\textsl{green}) and stationary estimates used as weak learners (\textsl{turquoise}). The orange curves correspond to the simulation used for Figures \ref{fig: sim_convergence} and \ref{fig: sim_whitening}.}
\label{fig: sim_sigmas}
 \end{figure}

The synthetic system is considered under two scenarios:
\begin{enumerate}[noitemsep] 
\item \textit{Nominal conditions}: when the system is stationary and no irregularities are present.
\item \textit{Disturbed conditions}: when the system is subject to a hypothetical irregularity in the center of the input domain.
\end{enumerate}
Testing our novel CBGP models under such nominal conditions is essential to ensure that they do not overfit or become unstable when nonstationarity is unnecessary. Tests under disturbed conditions are just as important to study how well irregularities can be accounted for. The hypothetical irregularity is injected into the domain under disturbed conditions so that the true signal and observation variations are given by the Hill function \citep{goutelle2008}
\begin{equation}
\label{eq: synthetic_sigma}
\sigma_\mathrm{X} = \sigma_{\mathrm{X},0} \left(
1 + \frac{c_\mathrm{X}}
{1 + \left(
    \frac{x - 2400}
    {360}
    \right)^6}
\right)
\end{equation}
where ``$\mathrm{X}$'' can be replaced with ``signal'' and ``obs'' and with $c_\mathrm{signal} = 5$ and $c_\mathrm{obs} = 3$. These inhomogeneous variation functions, represented by green curves in Figure \ref{fig: sim_sigmas}, describe a nonstationary GP model that otherwise assumes homogeneous correlation length scales that are unperturbed by the irregularity.

Our CBGP model is tested against the stationary GP model under both nominal and disturbed conditions. In the nominal scenario, the stationary model is exact and employs the true covariance and will outperform any other model but has no means to account for the irregularity under disturbed conditions other than $R_\mathrm{irreg}$ inflation post-fit. The CBGP will use the stationary covariance to serve as a weak learner for its boosting algorithm.

Truth consists of 5,000 simulations for both the nominal and disturbed conditions for this problem with a domain discretization size of 20 units. For each simulation, samplings consist of 30 simulated observations randomly chosen uniformly within the input domain to align with WAAS' target of 30 ionospheric pierce points for its local ionospheric fits \citet{sparks2011a}. All other parameters for the simulated testing scenario are provided in Appendix \ref{app: params}.

\begin{figure}[htb!]
 \centering
 \includegraphics[width=3.25in]{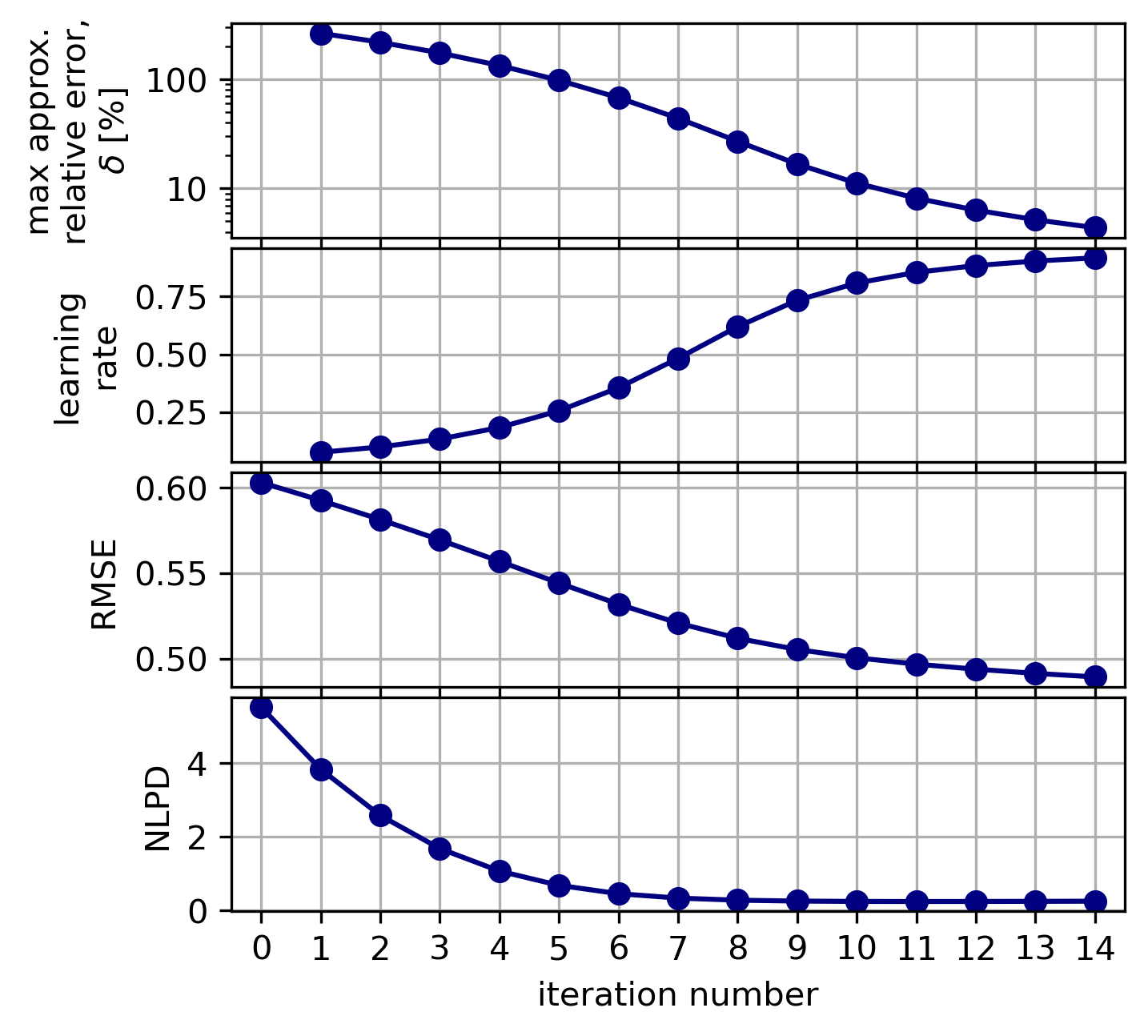}
\caption{Convergence and accuracy statistics of the iterative CBGP boosting scheme in Algorithm \ref{alg: cov_infl_alg} for a single one-dimensional simulated scenario under disturbed conditions.}
\label{fig: sim_convergence}
 \end{figure}

Convergence of the iterative boosting scheme for prior covariances outlined in Algorithm \ref{alg: cov_infl_alg} for one of the 5,000 simulations under disturbed conditions is demonstrated in Figure \ref{fig: sim_convergence}. The top panel of the figure shows an initial maximum relative error, $\delta$, of over 264\% for a stationary model at iteration \#0 to below a prescribed 5\% convergence tolerance at iteration \#12 specified by $\varepsilon=0.05$. The adaptive learning rate shown in the second figure panel increases as the relative error decreases and the scheme nears convergence. Root-mean-squared errors (RMSEs) and negative log predictive density (NLPD) in the third and fourth panels, respectively, demonstrate rapid accuracy improvements for posterior means and uncertainties from the stationary model in early iterations that begins to level off during the last few iterations.

\begin{figure}[htb!]
 \centering
 \includegraphics[width=6in]{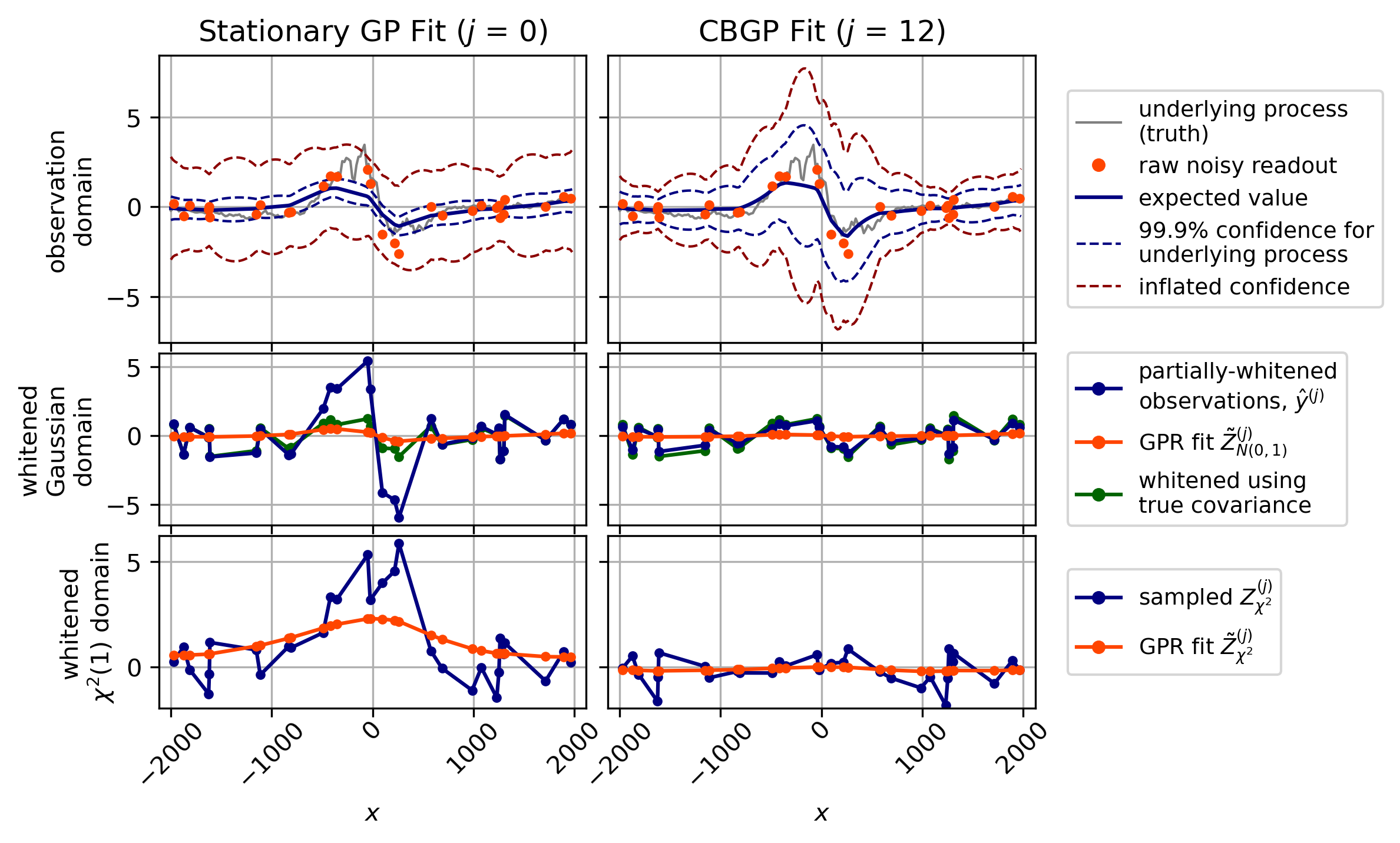}
\caption{GPR posteriors and whitened observations for stationary GP (\textit{left}) and CBGP (\textit{right}) fits for the single one-dimensional simulated scenario under disturbed conditions used for Figure \ref{fig: sim_convergence}. The top row of panels shows GPR posterior means and uncertainties fit to raw observations. The middle row shows partially-whitened observations (\textsl{navy}) along with the posterior mean of a GPR fit to them (\textsl{orange}) and observations whitened using the true covariance (\textsl{green}). The bottom row shows partially-whitened observations with the posterior mean of a GPR fit in the $\chi^2(1)$ domain.}
\label{fig: sim_whitening}
 \end{figure}

GPR fits for the simulated process used to generate Figure \ref{fig: sim_convergence} are shown in the top panel of Figure \ref{fig: sim_whitening} with stationary posteriors in the top left and CBGP posteriors shown in the top right. Truth and uncertainties shown do not account for observation noise, which is assumed to be included in the noisy readout of the sampling and only accounted for in the prior covariance. The stationary fit, whose signal and observation variations are underestimates where the irregularity is imposed, produces a mean, shown in navy, that does not capture fluctuations of the truth nearly as well as the CBGP fit's mean. Furthermore, the stationary's post-fit inflated uncertainty, quantified by $3.29 \sigma_\mathrm{infl}$ and represented by the dark red dashed line, does not bound all of the truth, represented in gray, failing to meet the three-nines requirement for this simulation as homogeneous $R_\mathrm{irreg}$ overinflates uncertainties near the boundaries of the domain where the process is less subject to the irregularity and is at more risk of underinflating near the irregularity in the center of the domain. The CBGP model, on the other hand, not only has an uninflated confidence interval, defined by $3.29 \sigma$ and represented by a dotted navy line, that is much greater where the irregularity is present but also produces a post-fit inflated confidence that is amplified in regions of the fit domain that are poorly sampled.

The second and third rows of Figure \ref{fig: sim_whitening} illustrate the intermediate GPR fitting of partially-whitened observations used for latent relative error estimation for the stationary initial state ($j=0$) and nonstationary state after $j=12$ iterations. The second row reveals how the iterative procedure refines initial partially-whitened observations, $\hat{y}^{(0)}$, that appear to have a relatively high autocorrelation to a seemingly uncorrelated state after 12 iterations, $\hat{y}^{(12)}$, that closely resembles the true whitened observations, shown in green. Similar behavior is observed in the last row of the figure where the partially-whitened observations mapped to a Gaussian distribution in the $\chi^2(1)$ domain, denoted by $\tilde{Z}_{\chi^2}^{(j)}$, lose their initial autocorrelation and ultimately resemble white noise. The GPR posterior means fit to the partially-whitened observations, $\tilde{Z}_{\mathcal{N}\left(0,1\right)}^{(j)}$ and $\tilde{Z}_{\chi^2}^{(j)}$ in the figure's second and third rows, respectively, mostly capture the morphology of the irregularity for the initial stationary case while nearly zero-valued on the entire domain after convergence of the iterative algorithm. The initial GPR fits that assume unit signal and observation noise for $j=0$ in the left column do not capture the entirety of the actual variation of the whitened observations in the Gaussian and $\chi^2(1)$ domains, but such underfitting is appropriate due to the nature of the boosting process that allows for uncaptured variation to be explained by further iterations with little overfitting risk.

Figure \ref{fig: sim_sigmas} shows the CBGP's latent signal and observation variation function estimates, $\sigma_\mathrm{signal}$ and $\sigma_\mathrm{obs}$, respectively, for the first 250 simulations under both nominal and disturbed conditions, shown in navy, along with the true and stationary initial $\sigma$'s, represented in green and turquoise, respectively. The orange curves shown on the right of the figure correspond to CBGP latent function estimates for the disturbed simulation highlighted in Figures \ref{fig: sim_convergence} and \ref{fig: sim_whitening}. For the nominal case, latent variation function estimates predominantly remain near the true stationary value and rarely exceed twice the true value. In disturbed settings, some estimated latent variation functions overestimate the true values at their peak while others underestimate, but, in general, the majority of the 250 capture features of the injected irregularity with inflated varation in the center of the domain that reduces outward toward the edges. However, nearly all disturbed simulations overapproximate near the edges of the domain to at least a slight extent due to a sharpness of the Hill function characterizing the irregularity in Equation \eqref{eq: synthetic_sigma} that may not be easily captured by the exponential decay kernel used for GP modeling of the partially-whitened observations while tuning $C^{(j)}$.  Nevertheless, this overestimation is not particularly egregious in most cases and, for the most part, general inhomogeneity stemming from the injected irregularity is successfully captured by the CBGP model. 

Computational results for the stationary GP and CBGP models aggregated across all 5,000 nominal and disturbed simulations are listed in Table \ref{table: sim_table}, which included 1,000,000 truth values for both sets of conditions. Included in the tables are RMSE, mean absolute error (MAE), NLPD, and Continuous Ranked Probability Score (CRPS) for $\mu$ and $\sigma$ with both NLPD and CRPS also computed using $\mu$ and $\sigma_\mathrm{infl}$. Under nominal conditions, the stationary model's exactness enable it to outperform the nonstationary model across all such metrics. However, the CBGP model is not drastically far off from the stationary model for the nominal scenario with aggregate RMSE and MAE that are only about 1\% higher than its stationary counterparts. As expected, the stationary model has almost exactly 99.9\% of its errors within 3.29$\sigma$ of the posterior mean while both stationary and CBGP models easily meet the three-nines requirement by having significantly more than 99.9\% of the errors within $\sigma_\mathrm{infl}$.

Under simulated disturbed conditions, discrepancies between the results for the stationary GP and CBGP models are magnified with the CBGP model providing substantially more accuracy. RMSEs and MAEs for CBGP errors under disturbed conditions, shown in the bottom row of Table \ref{table: sim_table}, exhibit approximately 33\% reductions from corresponding metric values for the stationary GP model. More notably, the CBGP model meets the three-nines integrity requirement but the stationary model does not with only about 99.41\% of its errors within $3.29\sigma_\mathrm{infl}$, falling short of the desired 99.9\% level. The inability of the stationary GP model to meet this requirement highlights the shortcomings of the $R_\mathrm{irreg}$ homogeneous inflation for GPs misspecified by inhomogeneous irregularities.

\begin{table}[htb!]
\begin{center}
\scriptsize
\begin{tabular}{ | c | c | c | c | c | c | c | c | c | c | }
\hline
\textbf{Conditions} & \textbf{Model} & \textbf{RMSE} & \textbf{MAE} & \textbf{NLPD} & \textbf{CRPS} & \makecell{\textbf{Within} \\ \textbf{$3.29\sigma$ [\%]} } & \makecell{\textbf{NLPD} \\ \textbf{(infl.)} } & \makecell{\textbf{CRPS} \\ \textbf{(infl.)} } & \makecell{\textbf{Within} \\ \textbf{$3.29\sigma_\mathrm{infl}$ [\%]} } \\ 
\hline
\hline
\multirow{2}{*}{nominal} & stationary GP & \textbf{0.181} & \textbf{0.142} & \textbf{-0.321} & \textbf{0.100} & 99.8979 & \textbf{-0.0913} & \textbf{0.111} & \textbf{99.9999} \\ 
\cline{2-10}
 & CBGP & 0.183 & 0.143 & -0.280 & 0.103 & 99.9617 & 0.0466 & 0.125 & \textbf{100} \\ 
\hline
\hline
\multirow{2}{*}{disturbed} & stationary GP & 0.710 & 0.427 & 6.92 & 0.363 & 77.8300 & 1.05 & 0.371 & 99.4108 \\ 
\cline{2-10}
 & CBGP & \textbf{0.535} & \textbf{0.321} & \textbf{0.427} & \textbf{0.236} & 98.7205 & \textbf{0.624} & \textbf{0.267} & \textbf{99.9195} \\ 
\hline
\end{tabular}
\end{center}
\caption{Aggregated error metrics from 5,000 one-dimensional simulated scenarios under both nominal (\textit{top}) and disturbed (\textit{bottom}) conditions.}
\label{table: sim_table}
\end{table}

\begin{figure}[htb!]
\centering
\includegraphics[width=5.75in]{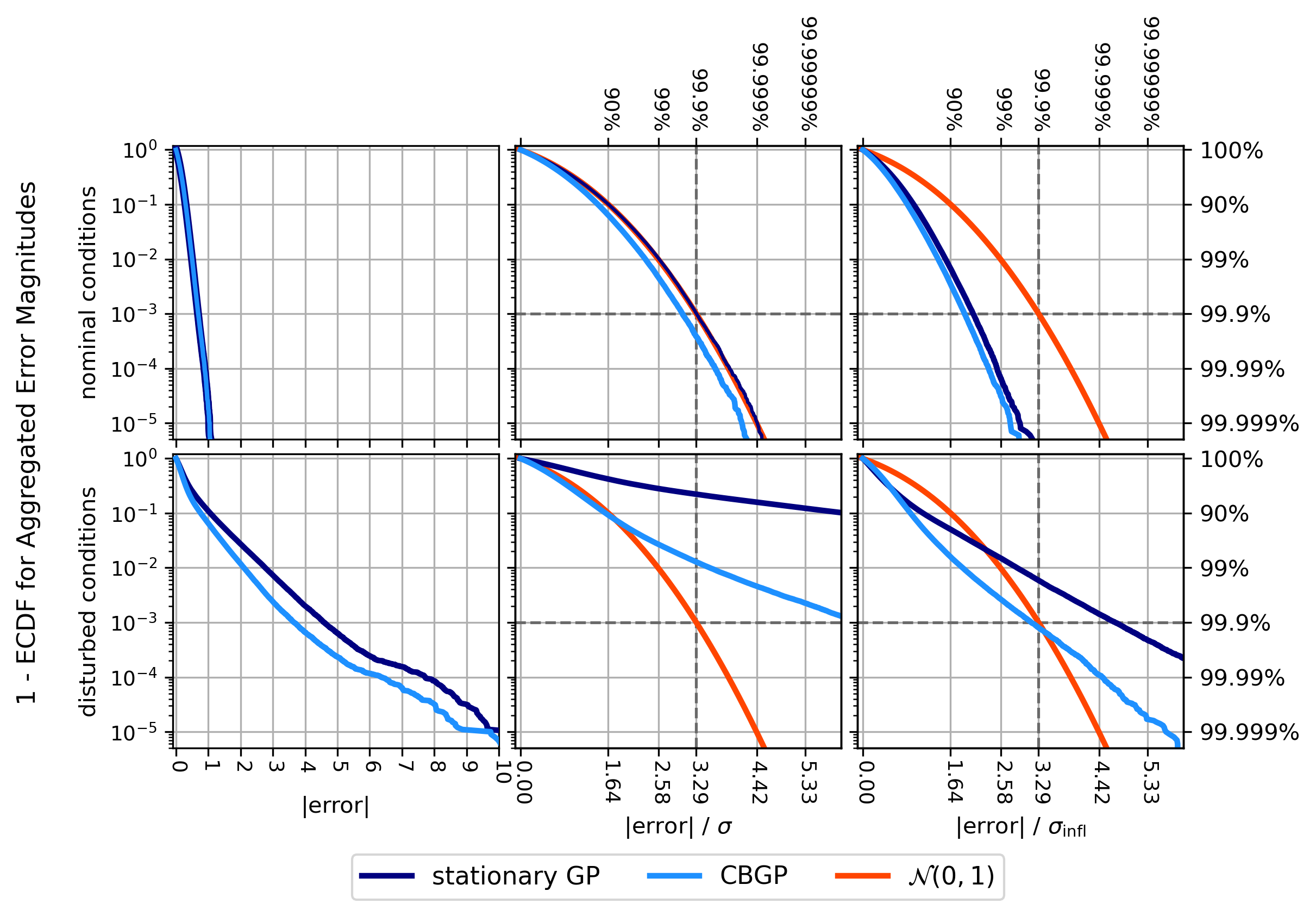}
\caption{Complementary ECDF curves for the magnitudes of out-of-sample errors (\textit{left}), errors standardized by the posterior uncertainty $\sigma$ (\textit{center}), and errors standardized by the inflated posterior uncertainty $\sigma_\mathrm{infl}$ (\textit{right}) aggregated from 5,000 one-dimensional simulated scenarios under both nominal (\textit{top}) and disturbed (\textit{bottom}) conditions.}
\label{fig: sim_ecdf}
 \end{figure}

Complementary empirical cumulative distribution function (ECDF) curves, quantified by $1-$ ECDF, for the magnitudes of stationary GP and CBGP model errors aggregated across all nominal and disturbed simulations displayed in Figure \ref{fig: sim_ecdf} further elucidate how the CBGP model properly handles both nominal and disturbed conditions while the stationary model struggles with the injected irregularity in disturbed conditions. The three columns of the figure show the complementary ECDF curves for the magnitudes of out-of-sample errors, errors standardized by the posterior uncertainty $\sigma$, and errors standardized by the inflated posterior uncertainty $\sigma_\mathrm{infl}$  in respective order for nominal conditions in the top row and disturbed conditions on the bottom row. Under nominal conditions, the error magnitude profile for the stationary model is nearly identical to that of the CBGP one and, when standardized by $\sigma$ aligns nearly perfectly with the standard unit normal curve, represented in orange. The CBGP model's complementary ECDF $\sigma$-standardized and $\sigma_\mathrm{infl}$-standardized curves sits just below their stationary counterparts due to slightly overinflated latent $\sigma_\mathrm{signal}$ and $\sigma_\mathrm{obs}$ estimates displayed in Figure \ref{fig: sim_sigmas}. Under disturbed conditions, though, differences in the models' complementary ECDF curves not only exhibit significantly improved accuracy by incorporating nonstationarity but also how the CBGP model meets the three-nines requirement while the stationary model is unable to. Complementary ECDF curves for disturbed $\sigma_\mathrm{infl}$-standarized errors for the stationary model rise above the standard unit normal curve before 2.58 is reached, suggesting that even a ``two-nines'' requirement is not met, while the CBGP curve lies below the standard unit normal one past 3.29.

\begin{figure}[htb!]
\centering
\includegraphics[width=4.5in]{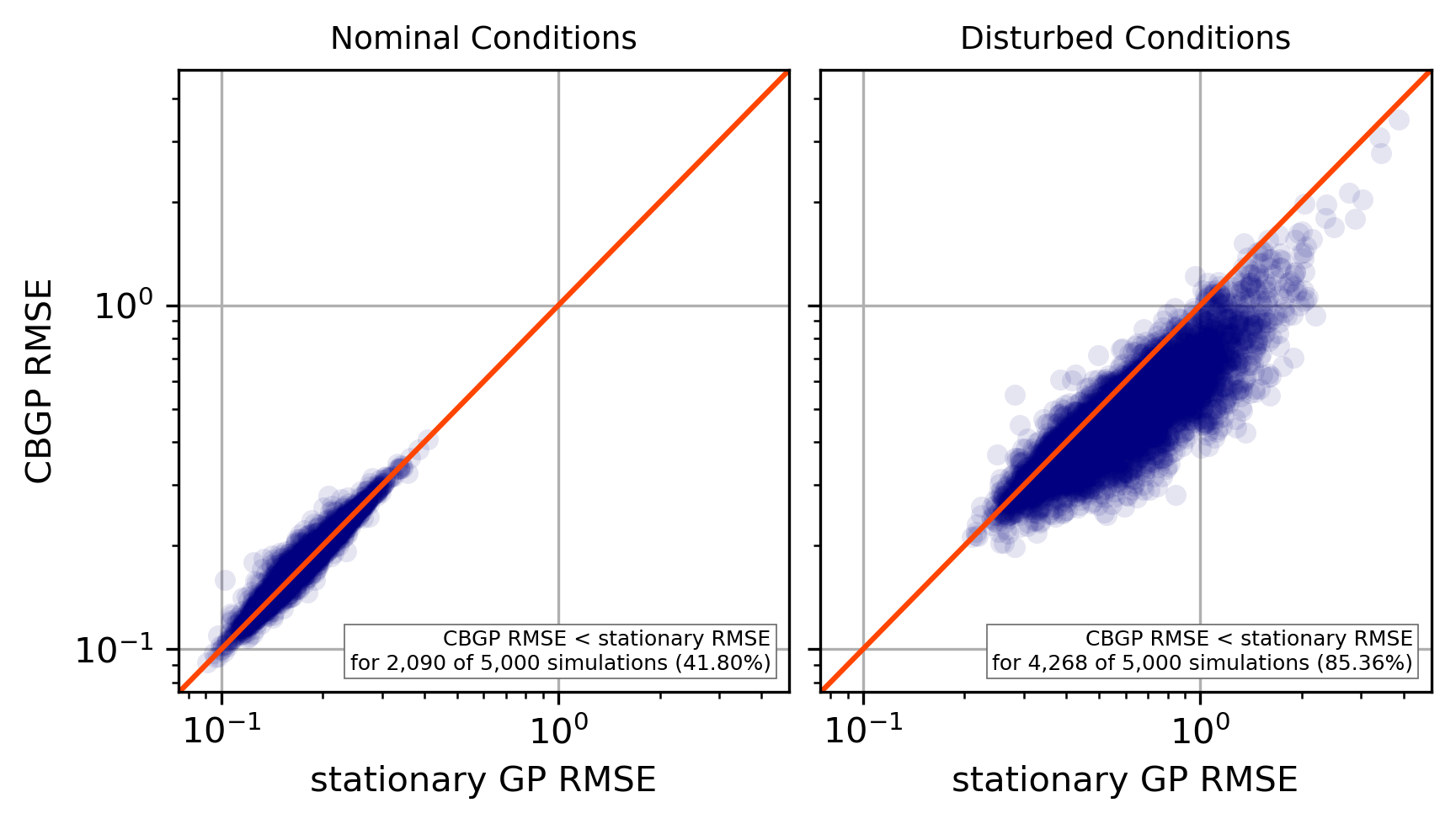}
\caption{Scatter plots comparing stationary GP and CBGP RMSEs across 5,000 one-dimensional simulations under nominal (\textit{left}) and disturbed (\textit{right}) conditions. The red curve represents the identity function, delineating RMSE equivalence for the two models.}
\label{fig: sim_scatter}
 \end{figure}

The accuracy of the CBGP fits relative to their stationary GP counterparts is evident in the scatter plot of their RMSEs for each of the 5,000 simulations under nominal and disturbed conditions shown in Figure \ref{fig: sim_scatter}. In the figure, the red line represents when the stationary and CBGP models achieve the same RMSE for a single simulation. Though the stationary GP captures the homogeneous system properties exactly, the CBGP still achieves a lower RMSE than the nonstationary model 41.80\% of the time under nominal conditions (2,010 of 5,000 instances), as captured by the proportion of markers below the red line in the left plot of the figure.  Under disturbed conditions though, the CBGP is more accurate than its stationary counterpart for 4,268 of the 5,000 simulations (85.36\%). Notably, for disturbed cases with stationary RMSEs exceeding a value of 1, the CBGP model has lower RMSEs 98.66\% of the time (589 of 597 instances).

This simulated testing case, which was constructed to be a dimensionally-reduced case of WAAS' ionospheric model under both nominal and disturbed conditions, provides encouraging results for CBGP fitting to actual ionospheric measurements in Section \ref{subsec: ionospheric_results}.


\subsection{Benchmark Experiments}
\label{subsec: benchmark_experiments}

To further test its efficacy, we evaluate the CBGP model against three other GP models for two benchmark experiments. Specifically, the CBGP model is compared with a stationary GP, a hetGPy implmentation of a heteroskedastic GP, and a nonstationary GP with a Gibbs-Paciorek kernel. Unlike the one-dimensional simulation tests in Section \ref{subsec: simulated_results}, the stationary GP will not be parameterized as the initial weak learner for the CBGP model  but instead using manually-selected parameterizations that better fit the observations.  All hetGPy tests use default parameterization settings with only the choice of kernel specified. Nonstationary Gibbs-Paciorek models with kernels described in Equation \eqref{eq: gibbs_paciorek} use latent functions for the length scale, signal variation, and observation variation determined by fitting Gaussian RBF basis functions by minimizing negative log marginal likelihoods. Due to the many degrees of freedom of having three different unknown inhomogeneous hyperparameters, initial guesses for Gibbs-Paciorek latent functions are given by the stationary GP model parameterizations. Both hetGPy and Gibbs-Paciorek implementations involve preprocessing by standardization using in-sample empirical means and standard deviations of both $x$ and $y$.

Parameterizations and computational details for all three comparison models are provided in Appendix \ref{app: params}.


\subsubsection{Motorcycle Acceleration Dataset Experiments}
\label{subsec: motorcycle_acceleration}

A seasoned litmus test for our CBGP framework is the modeling of simulated head acceleration during a motorcycle accident from \citet{silverman1985}, which has been used to benchmark numerous heterogeneous models \citep{kersting2007, heinonen2016, saul2016, binois2018, teemu2025} and as a featured application for hetGPy \citep{ogara2025}. However, to subject our framework to conditions for potential overconfidence and overfitting, we consider environments where about half of the measurements are deprived at a time while the other half are used for in-sample fitting by GP models. Specifically, we partition the input domain of the problem, $0 \leq t \leq 60$ ms, into the subset $M_1 = \left\{t \mid \mod(t + \frac{1}{2}\Delta t , 2\Delta t) \leq \Delta t \right\}$ and its complement $M_2 = M_1^\complement$ and fit each of the two subsets separately while depriving the other. Through this construction, interleaved intervals of size $\Delta t$ are withheld at a time for GPR fitting to impose significant gaps in sampling, rather than a uniform deprivation of observations throughout the fit domain. Larger $\Delta t$ values corresponding to larger gaps in sampling. For a given $\Delta t$, two experiments are performed: Experiment \#1 subsamples the measurement set $M_1$ while deprived of $M_2$ measurements, and Experiment \#2 subsamples $M_2$ while deprived of $M_1$. 

\begin{figure}[!htb]
\centering
\includegraphics[width=4.5in]{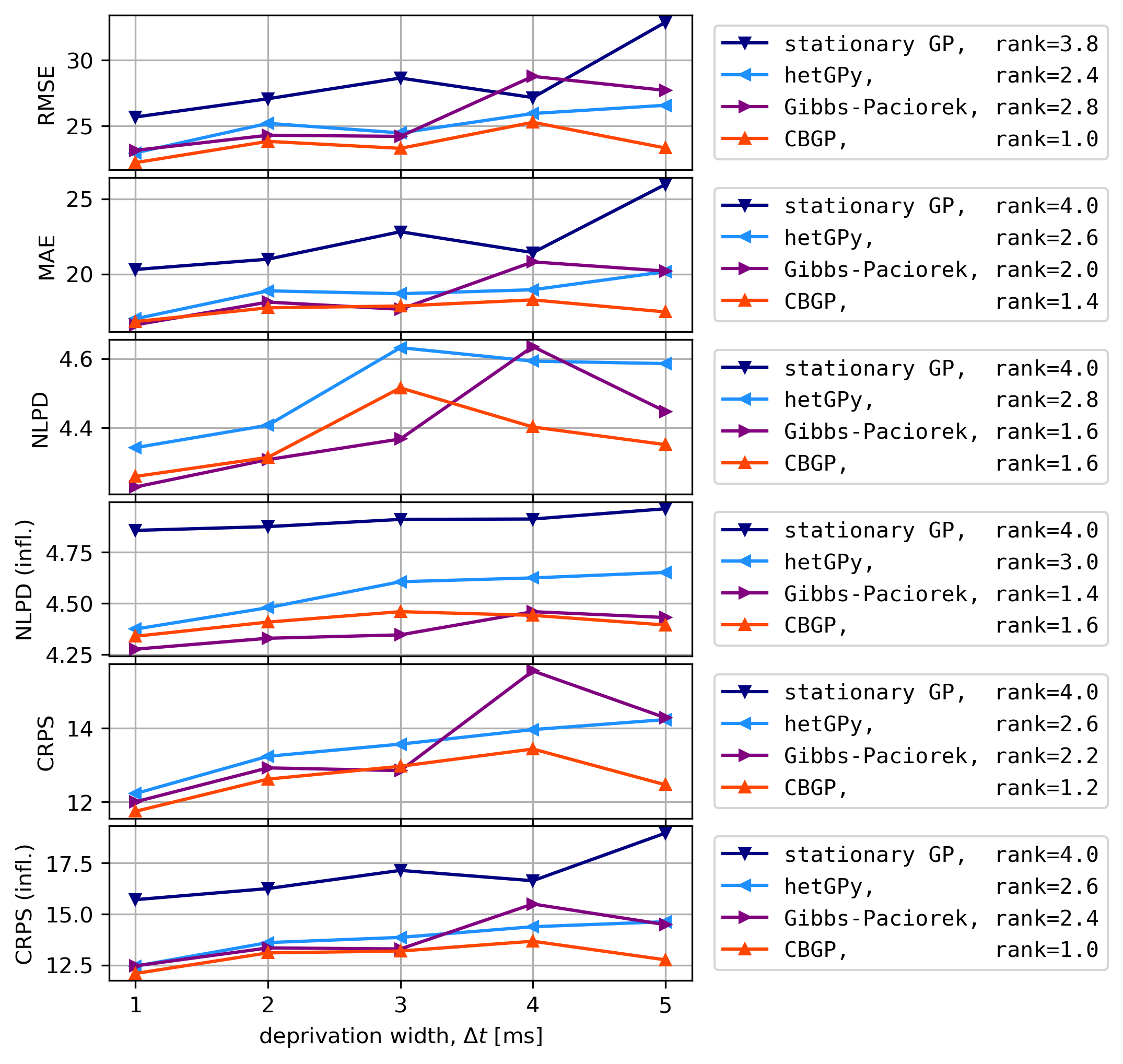}
\caption{Error statistics for Silverman's motorcycle accident dataset experiment for deprivation widths $\Delta t = 1, 2, 3, 4, 5$ ms along with average rank.}
\label{fig: mcycle_metrics}
\end{figure}

Error statistics and CPU times aggregated across Experiment \#1 and \#2 for deprivation widths $\Delta t = 1, 2, 3, 4, 5$ ms for all four GP models tested are listed in Table \ref{table: mcycle_table}. A visualization of error statistics and average rank across the four models is provided in Figure \ref{fig: mcycle_metrics} with rank values from 1 to 4 assigned to the lowest and highest values, respectively. First, all four of the models tested meet the three-nines requirement with $3.29\sigma_\mathrm{infl}$ bounding all out-of-sample, but the hetGPy implementation needs to use a Mat\'ern 3/2 kernel to achieve this requirement and is unable to do so using a Gaussian RBF kernel structure that is used by the other three models. In general, as the deprivation width $\Delta t$ increases, the error statistics for the models increase, suggesting poorer predictive power, as would be expected. The stationary GP has the lowest CPU times that are, on average, an order of magnitude lower than the other three inhomogeneous models, whose CPU times are mostly in-line with one another, but the stationary GP is, also, consistently outperformed by the other three models when using any of the error statistics studied. The CBGP ranks the highest of all four models for five of the six error statistics studied with the lone exception being NLPD computed using $\sigma_\mathrm{infl}$, for which the Gibbs-Paciorek model ranks highest. 

\begin{table}[htb!]
\begin{center}
\scriptsize
\begin{tabular}{ | c | c | c | c | c | c | c | c | c | }
\hline
\makecell{$\Delta t$ \\ \textbf{[ms]}} & \textbf{Model} & \makecell{\textbf{CPU} \\ \textbf{[s]} } & \makecell{\textbf{RMSE} \\ \textbf{[g]} } & \makecell{\textbf{MAE} \\ \textbf{[g]} } & \textbf{NLPD} & \makecell{\textbf{CRPS} \\ \textbf{[g]} } & \makecell{\textbf{NLPD} \\ \textbf{(infl.)} } & \makecell{\textbf{CRPS} \\ \textbf{(infl.) [g]} } \\ 
\hline
\hline
\multirow{4}{*}{1} & stationary GP & \textbf{0.0204} & 25.7 & 20.3 & 6.34 & 16.2 & 4.86 & 15.7 \\ 
\cline{2-9}
 & hetGPy & 0.116 & 23.0 & 17.0 & 4.34 & 12.2 & 4.37 & 12.4 \\ 
\cline{2-9}
 & Gibbs-Paciorek & 0.0983 & 23.2 & \textbf{16.6} & \textbf{4.23} & 12.0 & \textbf{4.28} & 12.5 \\ 
\cline{2-9}
 & CBGP & 0.134 & \textbf{22.2} & 16.8 & 4.26 & \textbf{11.7} & 4.34 & \textbf{12.1} \\ 
\hline
\hline
\multirow{4}{*}{2} & stationary GP & \textbf{0.00298} & 27.1 & 21.0 & 6.64 & 17.0 & 4.88 & 16.2 \\ 
\cline{2-9}
 & hetGPy & 0.219 & 25.2 & 18.9 & 4.41 & 13.2 & 4.48 & 13.6 \\ 
\cline{2-9}
 & Gibbs-Paciorek & 0.318 & 24.3 & 18.1 & \textbf{4.31} & 12.9 & \textbf{4.33} & 13.3 \\ 
\cline{2-9}
 & CBGP & 0.0881 & \textbf{23.8} & \textbf{17.8} & 4.31 & \textbf{12.6} & 4.41 & \textbf{13.1} \\ 
\hline
\hline
\multirow{4}{*}{3} & stationary GP & \textbf{0.00297} & 28.7 & 22.8 & 7.07 & 18.6 & 4.91 & 17.1 \\ 
\cline{2-9}
 & hetGPy & 0.183 & 24.5 & 18.7 & 4.63 & 13.6 & 4.61 & 13.9 \\ 
\cline{2-9}
 & Gibbs-Paciorek & 0.0866 & 24.2 & \textbf{17.7} & \textbf{4.37} & \textbf{12.9} & \textbf{4.35} & 13.3 \\ 
\cline{2-9}
 & CBGP & 0.0736 & \textbf{23.3} & 17.9 & 4.52 & 13.0 & 4.46 & \textbf{13.2} \\ 
\hline
\hline
\multirow{4}{*}{4} & stationary GP & \textbf{0.00200} & 27.2 & 21.4 & 6.66 & 17.2 & 4.91 & 16.6 \\ 
\cline{2-9}
 & hetGPy & 0.252 & 26.0 & 19.0 & 4.59 & 14.0 & 4.62 & 14.4 \\ 
\cline{2-9}
 & Gibbs-Paciorek & 0.110 & 28.8 & 20.8 & 4.63 & 15.6 & 4.46 & 15.5 \\ 
\cline{2-9}
 & CBGP & 0.0878 & \textbf{25.3} & \textbf{18.3} & \textbf{4.40} & \textbf{13.4} & \textbf{4.44} & \textbf{13.7} \\ 
\hline
\hline
\multirow{4}{*}{5} & stationary GP & \textbf{0.00199} & 32.9 & 26.0 & 8.21 & 21.5 & 4.96 & 19.0 \\ 
\cline{2-9}
 & hetGPy & 0.0949 & 26.6 & 20.2 & 4.59 & 14.2 & 4.65 & 14.6 \\ 
\cline{2-9}
 & Gibbs-Paciorek & 0.0990 & 27.7 & 20.2 & 4.45 & 14.3 & 4.43 & 14.5 \\ 
\cline{2-9}
 & CBGP & 0.0825 & \textbf{23.3} & \textbf{17.5} & \textbf{4.35} & \textbf{12.5} & \textbf{4.39} & \textbf{12.8} \\ 
\hline
\end{tabular}
\end{center}
\caption{Error statistics and CPU times for Silverman's motorcycle accident dataset experiment for deprivation widths $\Delta t = 1, 2, 3, 4, 5$ ms.}
\label{table: mcycle_table}
\end{table}

\begin{figure}[!htb]
 \centering
 \includegraphics[width=4.5in]{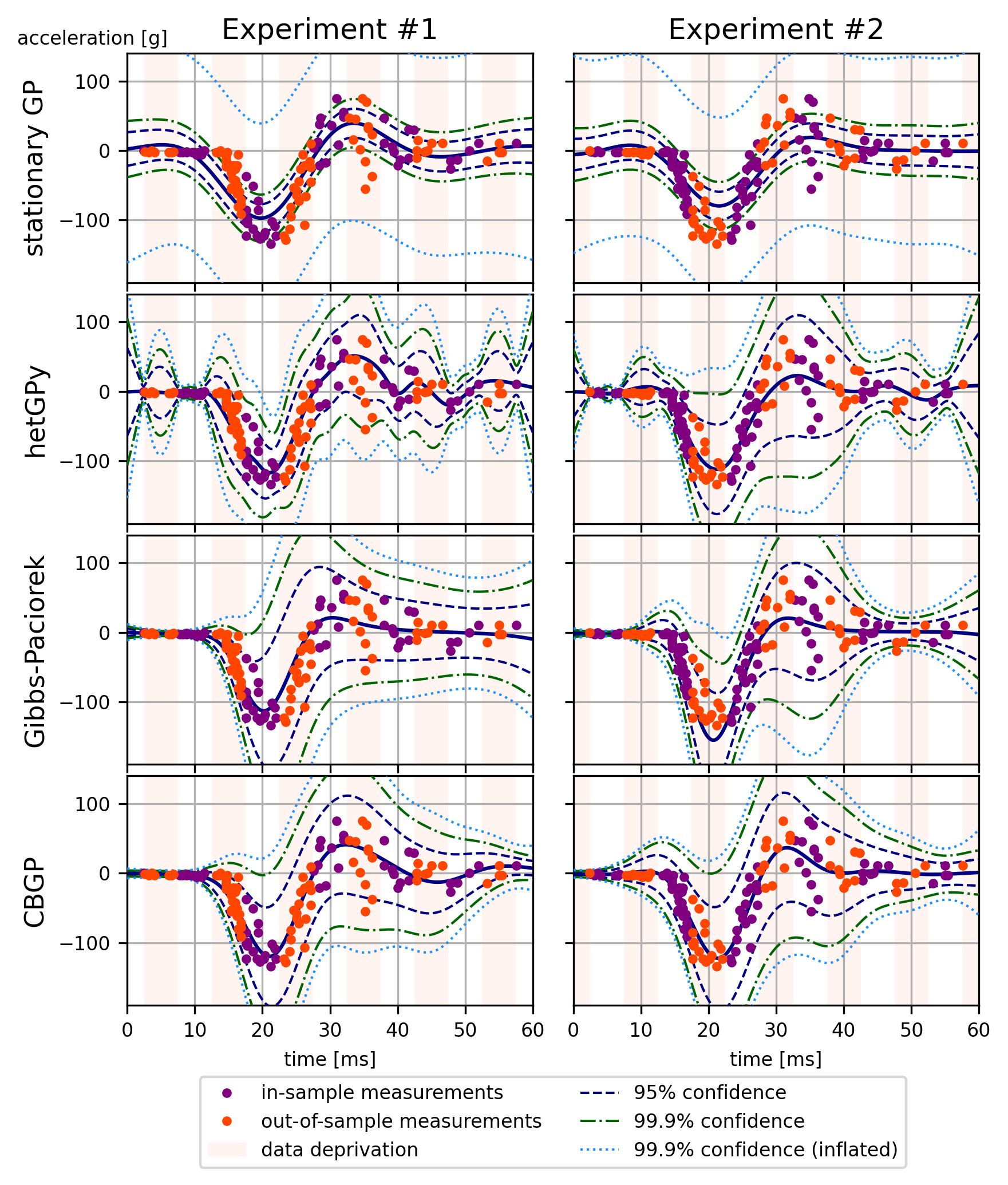}
\caption{GP posterior means and uncertainties for computational tests with two different data deprivation schemes, Experiment \#1 (\textit{left}) and \#2 (\textit{right}), applied to the Silverman's motorcycle accident dataset for the deprivation width $\Delta t = 5$ ms.}
\label{fig: mcycle_5}
 \end{figure}

Posterior distributions for two deprived testing cases with $\Delta t = 5$ ms are shown in Figure \ref{fig: mcycle_5}. In-sample measurements are depicted in purple while out-of-sample measurements are represented in orange with corresponding shading behind them denoting deprived subintervals. Posterior means for the GP models are shown in solid turquoise and navy curves, respectively. Confidence intervals for uncertainties without post-fit inflation that account for observation noise are represented by dotted lines at 95\% ($\mu\pm1.96\sigma$) and 99.9\% ($\mu\pm3.29\sigma$) along with a 99.9\% post-fit confidence interval ($\mu\pm3.29\sigma_\mathrm{infl}$). The CBGP modeling framework represents the ``nominal'' near-zero acceleration prior to the simulated accident around $t=13$ ms with an appropriately-small confidence interval and, after, captures the observed irregular variation during the accident with high integrity just as well or better than any of the other models tested. Across both Test \#1 and Test \#2, 126 of 133 total measurements (94.7\%) of the CBGP out-of-sample errors normalized by $\sigma$ are within the 95\% confidence interval, which increases to 131 measurements when normalized by $\sigma_\mathrm{infl}$ (98.5\%). Notably, the CBGP out-of-sample errors meet the requirement with and without post-fit inflation since all errors for the $\Delta t = 5$ ms scenario are within $3.29 \sigma$. Results from the two deprived testing cases show CBGP posterior distributions that fit both in-sample and out-of-sample measurements well but also meet integrity standards without overfitting nor overconfidence.

\subsubsection{Meuse River Dataset Experiments}
\label{subsec: real_estate_valuation}

Our second benchmark problem for the CBGP model consists of modeling five features of the Meuse River dataset \citep{rikken1993} in two spatial dimensions. This dataset consists of 155 measurements of concentrations of heavy metals in the topsoil near the Meuse River in the Netherlands along with the locations of the observations. Our testing problem consists of separately modeling the logartihm of the concentration of each of the four heavy metals, which are cadmium, copper, lead, and zinc, using a GP and also modeling the relative elevation of each sampling location as a GP in linear space. The concentrations of the heavy metals in the data set are correlated with the distance of the 155 sampling locations to the Meuse River and the relative elevation of each sampling but neither of these factors are accounted for in the GP models to prevent the inclusion of explanatory variables that reduce the inhomogeneity of GP modeling. Out-of-sample testing is conducted using a leave-one-out deprivation method for each of the observations for each of the five features. Similar to the hetGPy and Gibbs-Paciorek implementations, stationary GP and CBGP fits preprocess observations by standardization using in-sample empirical means and standard deviations.

Table \ref{table: meuse_table} displays CPU times and error statistics for computational experiments for the five Meuse River dataset features for each of the four GP models. Expected values for heavy metal GP RMSE and MAE statistics are evaluated using the expression $\mathbb{E}\left[y\right] = \exp\left(\mu + \frac{1}{2}\sigma^2\right)$ where $\mu$ and $\sigma$ are modeled in logarithmic space while all other statistics are computed in logarithmic space. As might be expected, the stationary GP computational times listed, which are aggregated for all 155 leave-one-out tests, are orders of magnitude smaller than the other three inhomogeneous models. The CBGP is the next most efficient model in terms of CPU usage with times that are on average at least half of the corresponding times for the hetGPy and Gibbs-Paciorek implementations. No single model is the most accurate in describing leave-one-out errors for the Meuse River dataset with each of the models achieving the lowest RMSE for at least one of the five features: the stationary GP's cadmium RMSE is lowest; hetGPy's elevation RMSE is lowest; the Gibb-Paciorek's zinc RMSE is lowest; and the CBGP's copper and lead RMSEs are lowest. However, the CBGP is consistently relatively accurate with the lowest or second lowest RMSE, MAE, NLPD, and CRPS for all five features modeled. Furthermore, the CBGP model achieves the lowest NLPD values for all four heavy metal concentrations while still bounding the largest proportion of errors within the uninflated confidence interval specified by $3.29\sigma$ for four of the five features.

One important observation is that the stationary GP and CBGP models successfully meet the three-nines integrity requirement with post-fit inflated uncertainties, as evidenced by corresponding percentages exceeding 99.9\% in the last column of Table \ref{table: meuse_table}, but the hetGPy and Gibbs-Paciorek tests fail to do so for all but one test. This outcome suggests that, for the Meuse River dataset, $R_\mathrm{irreg}$ inflation is insufficient in properly bounding residuals for inhomogeneous hetGPy and Gibbs-Paciorek implementations. One tradeoff for the CBGP meeting the integrity requirement is higher $\sigma_\mathrm{infl}$ values that resulted in higher NLPD and CRPS values computed using $\sigma_\mathrm{infl}$, as shown in the third- and second-to-last columns in Table \ref{table: meuse_table}, respectively.

\begin{table}[htb!]
\begin{center}
\scriptsize
\begin{tabular}{ | c | c | c | c | c | c | c | c | c | c | c | }
\hline
\textbf{Feature} & \textbf{Model} & \textbf{CPU [s]} & \textbf{RMSE} & \textbf{MAE} & \textbf{NLPD} & \textbf{CRPS} & \makecell{\textbf{Within} \\ \textbf{$3.29\sigma$ [\%]} } & \makecell{\textbf{NLPD} \\ \textbf{(infl.)} } & \makecell{\textbf{CRPS} \\ \textbf{(infl.)} } & \makecell{\textbf{Within} \\ \textbf{$3.29\sigma_\mathrm{infl}$ [\%]} } \\ 
\hline
\hline
\multirow{4}{*}{cadmium} & stationary GP & \textbf{0.368} & \textbf{2.87} & \textbf{1.65} & 3.57 & 0.551 & 79.355 & 1.38 & 0.515 & \textbf{100} \\ 
\cline{2-11}
 & hetGPy & 127. & 5.44 & 2.50 & 1.29 & 0.499 & 97.419 & 1.29 & 0.507 & 99.355 \\ 
\cline{2-11}
 & Gibbs-Paciorek & 66.0 & 4.55 & 2.15 & 1.29 & \textbf{0.458} & 98.065 & \textbf{1.24} & \textbf{0.465} & 98.710 \\ 
\cline{2-11}
 & CBGP & 37.4 & 3.05 & 1.88 & \textbf{1.27} & 0.463 & 98.710 & 1.45 & 0.584 & \textbf{100} \\ 
\hline
\hline
\multirow{4}{*}{copper} & stationary GP & \textbf{0.353} & 17.2 & 12.2 & 2.04 & 0.223 & 79.355 & 0.415 & 0.199 & \textbf{100} \\ 
\cline{2-11}
 & hetGPy & 102. & 15.5 & 10.8 & 0.370 & 0.177 & 98.710 & 0.367 & 0.180 & 98.710 \\ 
\cline{2-11}
 & Gibbs-Paciorek & 111. & 15.5 & 10.7 & 0.301 & 0.175 & 98.065 & \textbf{0.239} & \textbf{0.176} & 99.355 \\ 
\cline{2-11}
 & CBGP & 35.4 & \textbf{14.1} & \textbf{10.0} & \textbf{0.278} & \textbf{0.172} & 98.065 & 0.525 & 0.210 & \textbf{100} \\ 
\hline
\hline
\multirow{4}{*}{lead} & stationary GP & \textbf{0.356} & 78.0 & 49.8 & 1.78 & 0.247 & 88.387 & 0.601 & 0.233 & \textbf{100} \\ 
\cline{2-11}
 & hetGPy & 124. & 105. & 58.9 & 0.635 & 0.235 & 98.065 & 0.598 & 0.241 & 99.355 \\ 
\cline{2-11}
 & Gibbs-Paciorek & 101. & 75.8 & 49.7 & 0.594 & 0.225 & 97.419 & \textbf{0.516} & \textbf{0.227} & 98.710 \\ 
\cline{2-11}
 & CBGP & 37.2 & \textbf{74.1} & \textbf{47.4} & \textbf{0.455} & \textbf{0.213} & 98.710 & 0.684 & 0.268 & \textbf{100} \\ 
\hline
\hline
\multirow{4}{*}{zinc} & stationary GP & \textbf{0.361} & 250. & 158. & 1.66 & 0.258 & 89.032 & 0.651 & 0.246 & \textbf{100} \\ 
\cline{2-11}
 & hetGPy & 123. & 270. & 156. & 0.583 & 0.227 & 98.065 & 0.553 & 0.231 & 98.710 \\ 
\cline{2-11}
 & Gibbs-Paciorek & 77.6 & \textbf{214.} & 136. & 0.469 & \textbf{0.207} & 98.065 & \textbf{0.437} & \textbf{0.209} & \textbf{100} \\ 
\cline{2-11}
 & CBGP & 37.4 & 217. & \textbf{136.} & \textbf{0.461} & 0.209 & 97.419 & 0.675 & 0.269 & \textbf{100} \\ 
\hline
\hline
\multirow{4}{*}{elevation} & stationary GP & \textbf{0.342} & 0.785 & 0.624 & 3.53 & 0.502 & 77.419 & 1.24 & 0.451 & \textbf{100} \\ 
\cline{2-11}
 & hetGPy & 133. & \textbf{0.767} & 0.608 & 1.29 & 0.438 & 98.065 & 1.22 & 0.442 & 98.710 \\ 
\cline{2-11}
 & Gibbs-Paciorek & 59.8 & 0.779 & \textbf{0.595} & \textbf{1.22} & \textbf{0.434} & 98.710 & \textbf{1.21} & \textbf{0.440} & 99.355 \\ 
\cline{2-11}
 & CBGP & 32.2 & 0.778 & 0.598 & 1.25 & 0.436 & 98.710 & 1.47 & 0.536 & \textbf{100} \\ 
\hline
\end{tabular}
\end{center}
\caption{Error statistics and CPU times for the Meuse River dataset experiment.}
\label{table: meuse_table}
\end{table}

\begin{figure}[!htb]
 \centering
 \includegraphics[width=5in]{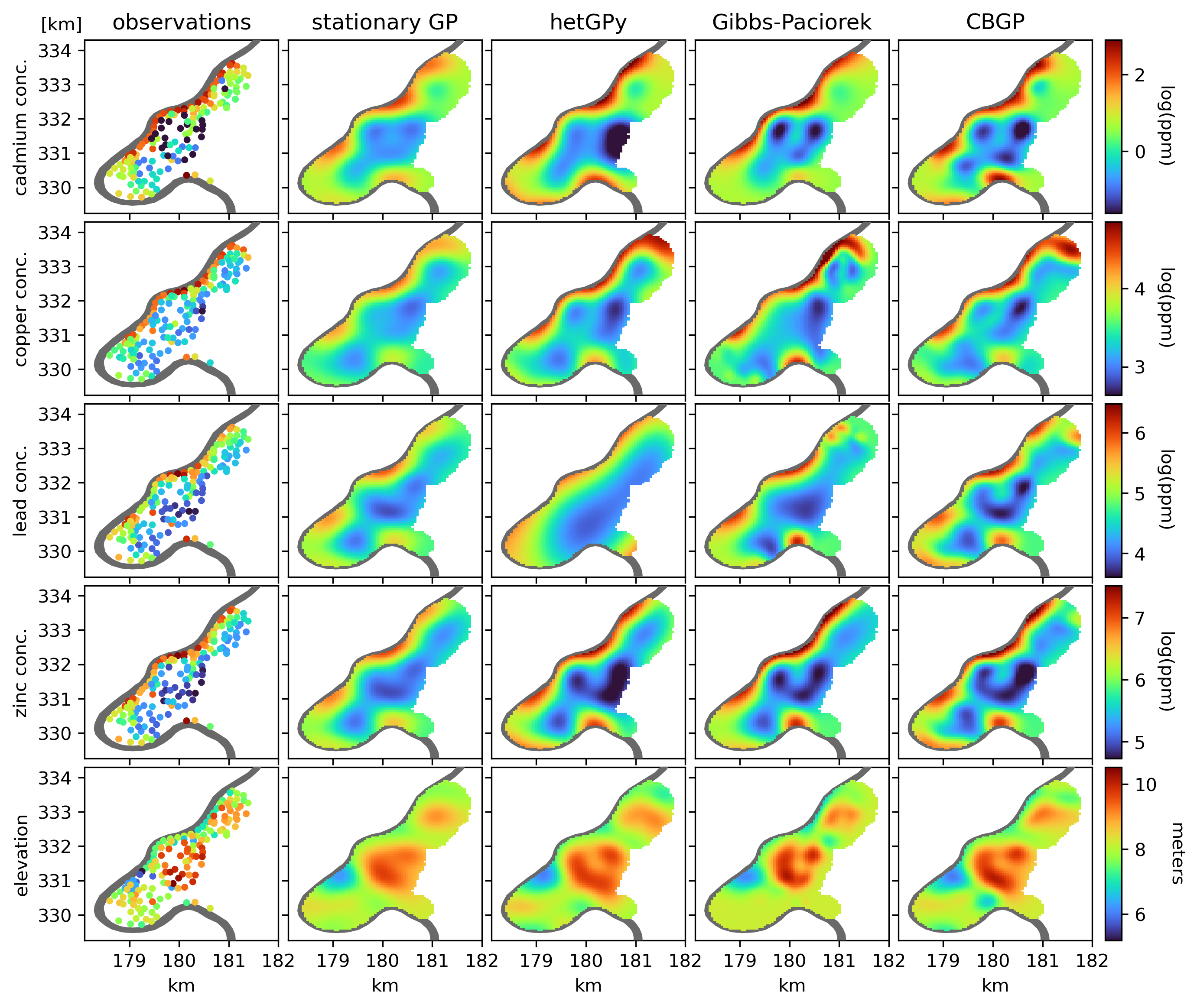}
\caption{GP posterior means for five features from the Meuse River dataset. Concentrations for the four minerals are shown in logarithmic space.}
\label{fig: meuse_posteriors}
\end{figure}

A visualization of posterior means fit to all observations for each of the five features and for all four GP models is shown in Figure \ref{fig: meuse_posteriors}. In general, the three inhomogeneous GPs capture the high density of heavy metals near the river bank better than the stationary model. The Gibbs-Paciorek GP and the CBGP appear to better resolve differences between observation noise and signal variation than hetGPy. For example, hetGPy attributes the vast majority of the variability of the lead concentration measurements to heteroskedastic observation noise and uses a long length scale to describe the underlying GP. The Gibbs-Paciorek and CBGP posterior means compare well to one another; however, at times, the latent Gibbs-Paciorek length scales appear to be quite small for copper and lead concentrations near the spatial coordinates $(181, 333)$ and seemingly result in minor overfitting nearby.

\subsection{Ionospheric Modeling}
\label{subsec: ionospheric_results}

This section focuses on the application that motivated the development of CBGP methods in this paper: ionospheric modeling for SBAS. We test our algorithms for the ionospheric region covering South America, which is subject to the Equatorial Ionization Anomaly (EIA) \citep{balan2018} and South Atlantic geomagnetic anomaly \citep{abdu2005}, where space weather is the most active globally. South America is not currently covered by an SBAS with testing of WAAS-like algorithms suggesting severely limited availability there \citep{lejeune2003}.

This application to the ionosphere differs from the previous computational tests in this paper in the sense that CBGP modeling is performed both locally through subsampling and using all measurements within the spatial domain of observations. In particular, we test the following three GP models:
\begin{enumerate}[noitemsep] 
\item WAAS' stationary GP model with linear drift outlined by \citet{sparks2011a} that fits ionospheric measurements in a local neighborhood.
\item The CBGP model with linear drift that fits ionospheric measurements in a local neighborhood. 
\item The CBGP model with de-meaning that fits all available ionospheric measurements over the entire test region at once.
\end{enumerate}
Linear drift modeling and de-meaning is described in Appendix \ref{app: incorporating_linear_drift}, and CBGP methods in Algorithm \ref{alg: cov_infl_alg} are assumed to be modified accordingly for this application. For each of the two nonstationary GP models, the ``nominal'' prior covariance parameterization listed in Table \ref{table: CBGP_params} is identical to WAAS' parameterization \citet{sparks2011a}. Following all other non-NCGBP tests in this paper and WAAS' convention, $R_\mathrm{irreg}$ inflation is assumed for the stationary model.

To summarize the WAAS model, ionospheric vertical total electronic content (TEC) measurements collected simultaneously at a single instance in time, or epoch, are fit using a spatial GP with an underlying planar model describing local behavior about a single ionospheric grid point (IGP) from a set of many IGPs that cover a region of interest. First, the ionosphere is assumed to be reduced to a single shell and ionospheric pierce points (IPPs) are determined, which specify a single location where all ionospheric content causing signal delay is assumed to reside. Then, the location of all measured IPPs are translated to an East-North-Up coordinate system relative to the IGP where the fit will be performed. For a single fit about an IGP, IPPs are subsampled to a local GPR fit neighborhood within a certain distance from the IGP within the east-north plane. This distance may range from 800 km to 2,100 km and is determined using an algorithm that targets 30 IPPs with measurements for fitting but still allows to use as few as 10 IPPs \citep{sparks2011a}. Slant delay measurements collected along the satellite-to-receiver ray path are translated to the vertical domain specified to be orthogonal to the Earth's surface at the IPP using a mapping function \citep{ovadia2025}. WAAS models are evaluated using the linear drift approach in Appendix \ref{app: incorporating_linear_drift} using a $G$ matrix determined using only east and north IPP coordinates relative to the IGP. WAAS' GP models are evaluated with a stationary exponential decay kernel and have been shown to be more accurate than planar fits alone \citep{sparks2011a}.

\begin{figure}[!htb]
 \centering
 \includegraphics[width=2.75in]{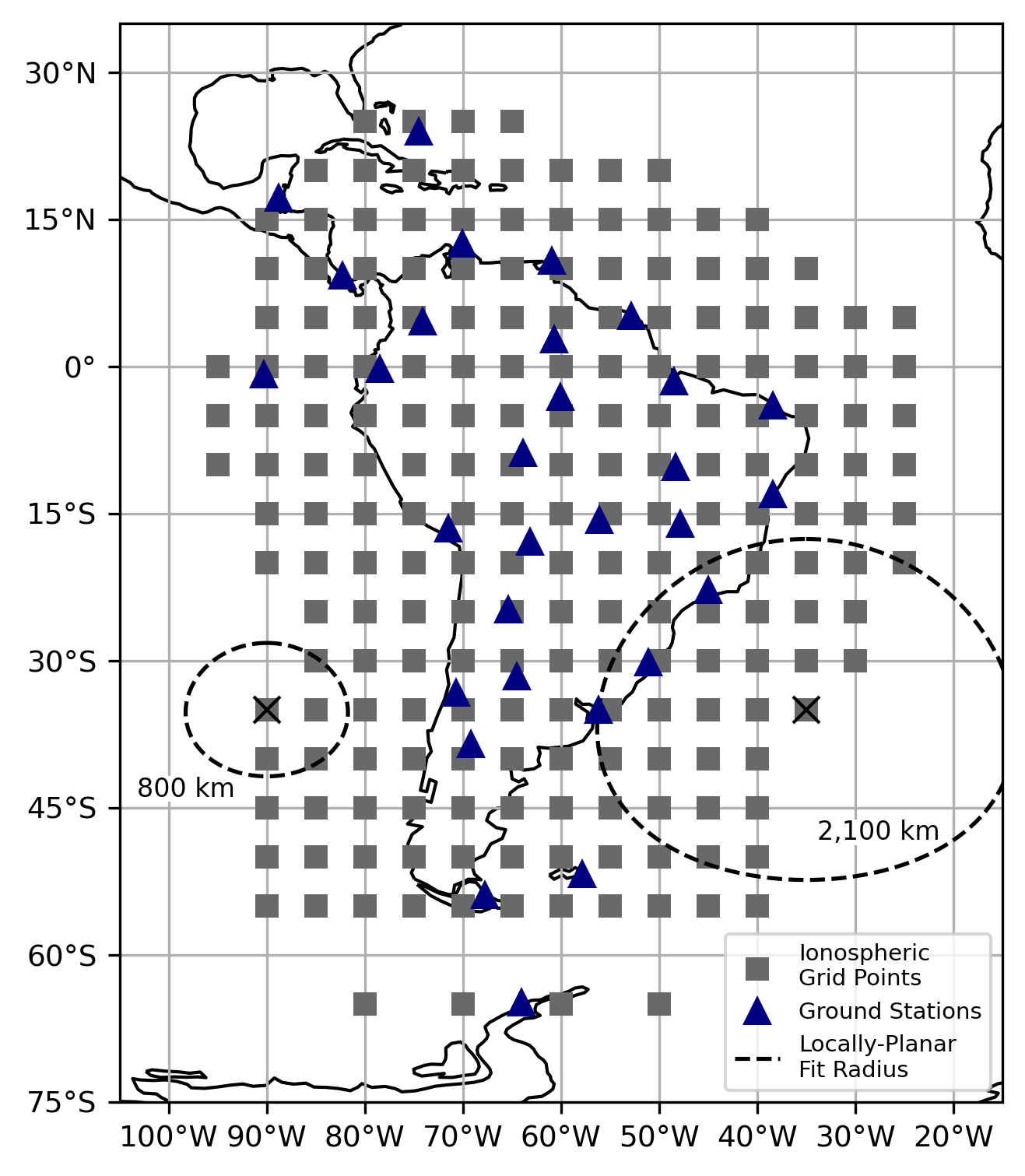}
\caption{Set of Ionospheric Grid Points (IGPs) with GNSS receiver sites used for ionospheric measurements. Dotted lines represent different potential sample neighborhoods for locally planar fits about a single IGP, represented with an ``X'' marker.}
\label{fig: station_and_IGP_set}
 \end{figure}

The set of IGPs used for the study is determined as the subset from those specified in the Minimum Operational Performance Standards (MOPS) IGP mask \citep{rtca2020} that encapsulates all lines-of-sight visible from mainland South America and the Falkand Islands with elevation angles 15$^\circ$ and above, as represented by grey square markers in Figure \ref{fig: station_and_IGP_set}. The dotted curves surrounding IGPs denoted by ``X'' markers at (35$^\circ$S, 90$^\circ$W) and (35$^\circ$S, 35$^\circ$W) in the figure illustrate potential 800 km and 2,100 km neighborhoods, respectively, used for fitting by local GP models with linear drift. For locally planar stationary and nonstationary GPR fits at those two IGPs with corresponding neighborhoods, observations for fitting would be subsampled to include only IPPs encapsulated by the dotted lines.

The locally planar CBGP model is evaluated similar to WAAS' model except with a linear drift version of Algorithm \ref{alg: cov_infl_alg} replacing a stationary GP with linear drift; however, the regional CBGP model for our tests is much different. When it converges in a single iteration, the locally planar CBGP will yield posterior $\mu$ and $\sigma$ values that are nearly identical to those from WAAS' model, but its inflated uncertainty, $\sigma_\mathrm{infl}$, will be notably different since it does not employ WAAS' $R_\mathrm{irreg}$ inflation. On the other hand, regional CBGP model posteriors will be significantly different than its WAAS counterparts since it deviates from typical SBAS conventions of linearizing the ionosphere locally and does not subsample measurements collected. Instead, planar assumptions are removed for the regional model, and all observations available are used for fitting. In this scenario, a constant drift model from Appendix \ref{app: incorporating_linear_drift} de-means these observations in a BLUP fashion using $G=\vec{1}$. Such an approach departs from planar assumptions adopted by, not only WAAS, but other SBAS, including the MTSAT-based Satellite Augmentation System (MSAS) \citep{sakai2017}, the European Geostationary Navigation Overlay System (EGNOS) \citep{olabode2022}, and the GPS Aided Geo Augmented Navigation (GAGAN) system \citep{sarma2009}.

Absolute ionospheric measurements, considered as truth, are sourced from 30 Global navigation satellite system (GNSS) receivers that provide observables to the International GNSS Service (IGS) \citep{johnston2017} and the University Navstar Consortium (UNAVCO) located in and around South America. Truth is computed from GPS and Galileo satellite constellations using computational methods described by \citet{ovadia2025} with an assumed ionospheric shell height of 450 km. GNSS receiver information is provided in Table \ref{table: iono_stations} and are represented as navy trianglies in the map shown in Figure \ref{fig: station_and_IGP_set}. Station selection was determined manually by potential availability of measurements from both GPS and Galileo satellite constellations and by spatial proximity to create a nearest-neighbor separation distance on the order of hundreds of kilometers, which is typically used for SBAS ground networks.

\begin{table}[htb!]
\begin{center}
\scriptsize
\begin{tabular}{ |c|c|c|c|c|c| }
\hline
\textbf{Station} & \textbf{Network} & \textbf{Latitude [$^{\circ}$]} & \textbf{Longitude [$^{\circ}$]} & \textbf{Altitude [m]} & \textbf{Constellations} \\
\hline
AREG & IGS & -16.47 & -71.49 & 2488.9 & GPS, Galileo \\
BELE & IGS & -1.41 & -48.46 & 9.1 & GPS \\
BOAV & IGS & 2.85 & -60.7 & 69.5 & GPS, Galileo \\
BOGT & IGS & 4.64 & -74.08 & 2,576.8 & GPS, Galileo \\
BRAZ & IGS & -15.95 & -47.88 & 1106 & GPS, Galileo \\
BRFT & IGS & -3.88 & -38.43 & 21.7 & GPS, Galileo \\
CHPI & IGS & -22.69 & -44.99 & 617.5 & GPS, Galileo \\
CN13 & UNAVCO & 24.07 & -74.53 & -34.5 & GPS, Galileo \\
CN19 & UNAVCO & 12.61 & -70.05 & -0.4 & GPS, Galileo \\
CN20 & UNAVCO & 9.35 & -82.26 & 13.2 & GPS, Galileo \\
CN23 & UNAVCO & 17.26 & -88.78 & 55.3 & GPS, Galileo \\
CN57 & UNAVCO & 10.84 & -60.94 & -5.6 & GPS, Galileo \\
CORD & IGS & -31.53 & -64.47 & 747.1 & GPS, Galileo \\
CUIB & IGS & -15.56 & -56.07 & 237.4 & GPS, Galileo \\
FALK & IGS & -51.69 & -57.87 & 50.8 & GPS, Galileo \\
GLPS & IGS & -0.74 & -90.3 & 1.8 & GPS, Galileo \\
KOUR & IGS & 5.25 & -52.81 & -25.8 & GPS, Galileo \\
MTV2 & IGS & -34.91 & -56.18 & 40.8 & GPS \\
NAUS & IGS & -3.02 & -60.06 & 93.9 & GPS, Galileo \\
PALM & IGS & -64.78 & -64.05 & 31.2 & GPS, Galileo \\
POAL & IGS & -30.07 & -51.12 & 76.7 & GPS, Galileo \\
POVE & IGS & -8.71 & -63.9 & 119.6 & GPS, Galileo \\
PWRO & UNAVCO & -38.42 & -69.21 & 491.6 & GPS \\
QUI3 & IGS & -0.14 & -78.47 & 2,927.5 & GPS \\
RGDG & IGS & -53.79 & -67.75 & 32.4 & GPS, Galileo \\
SANT & IGS & -33.15 & -70.67 & 723.1 & GPS, Galileo \\
SAVO & IGS & -12.94 & -38.43 & 76.3 & GPS, Galileo \\
SCRZ & IGS & -17.8 & -63.16 & 438.3 & GPS, Galileo \\
TOPL & IGS & -10.17 & -48.33 & 256.6 & GPS, Galileo \\
UNSA & IGS & -24.73 & -65.41 & 1257.8 & GPS, Galileo \\
\hline
\end{tabular}
\end{center}
\caption{GNSS receiver sites providing measurements used to compute ionospheric truth.}
\label{table: iono_stations}
\end{table}

For breadth and intensiveness, our computational tests are conducted on 27,944,482 ionospheric measurements collected over four of South America's most disturbed five-day space weather events since 2023. Each of the storms and their characterization is listed in Table \ref{table: iono_storm_set}. The March 2023 and 2025 storms are selected due to the presence of steep spatial ionospheric gradient magnitudes, as identified by rates of change of global ionospheric maps published by the Jet Propulsion Laboratory \citep{martire2024} computed using second-order central difference approximations. The May and October 2024 storms are selected due to the presence of extreme geomagnetic activity. These two 2024 timeframes coincide with notably high Planetary-K (Kp) index values, as computed by GFZ Helmholtz Centre Potsdam \citep{matzka2021}, that quantifies disturbances to the Earth’s geomagnetic field and Disturbance Storm Time (Dst) index values, as computed by Kyoto University \citep{takahashi1990}, that quantify the intensity of the symmetrical equatorial electrojet across the globe. All GPS and Galileo measurements collected above a 5$^\circ$ line-of-sight elevation mask from the 30 GNSS receivers during these storms are included at a sampling rate of 30 seconds/sample, except for manually detected cycle slip events that yield erroneous measurements and prompt removal from the data set. Overall, our set of observations consist of 57,600 distinct epochs for GPR fitting with locally planar GP model evaluations performed at as many of the 206 IGPs in Figure \ref{fig: station_and_IGP_set} as possible for every epoch and the regional CBGP regional model only performing one per epoch. Due to diurnal fluctuations of the ionosphere, some nighttime epochs may correspond to nominal or quiet ionospheric conditions while daytime epochs often occur during highly active or disturbed conditions during ionospheric storms.

\begin{table}[htb!]
\begin{center}
\footnotesize
\begin{tabular}{ |c|c|c|c| }
\hline
\textbf{Start Date} & \textbf{End Date} & \textbf{Characterized By} &  \makecell{\textbf{Receivers Without} \\ \textbf{Measurements} } \\
\hline
2023-03-11 & 2023-03-15 & Large Ionospheric Gradients & BOAV  \\
2024-05-09 & 2024-05-13 & High Geomagnetic Activity & NAUS  \\
2024-10-08 & 2024-10-12 & High Geomagnetic Activity &  - \\
2025-03-24 & 2025-03-28 & Large Ionospheric Gradients & -  \\
\hline
\end{tabular}
\end{center}
\caption{Storm set used for ionospheric model testing.}
\label{table: iono_storm_set}
\end{table}

\begin{figure}[!htb]
 \centering
 \includegraphics[width=5.5in]{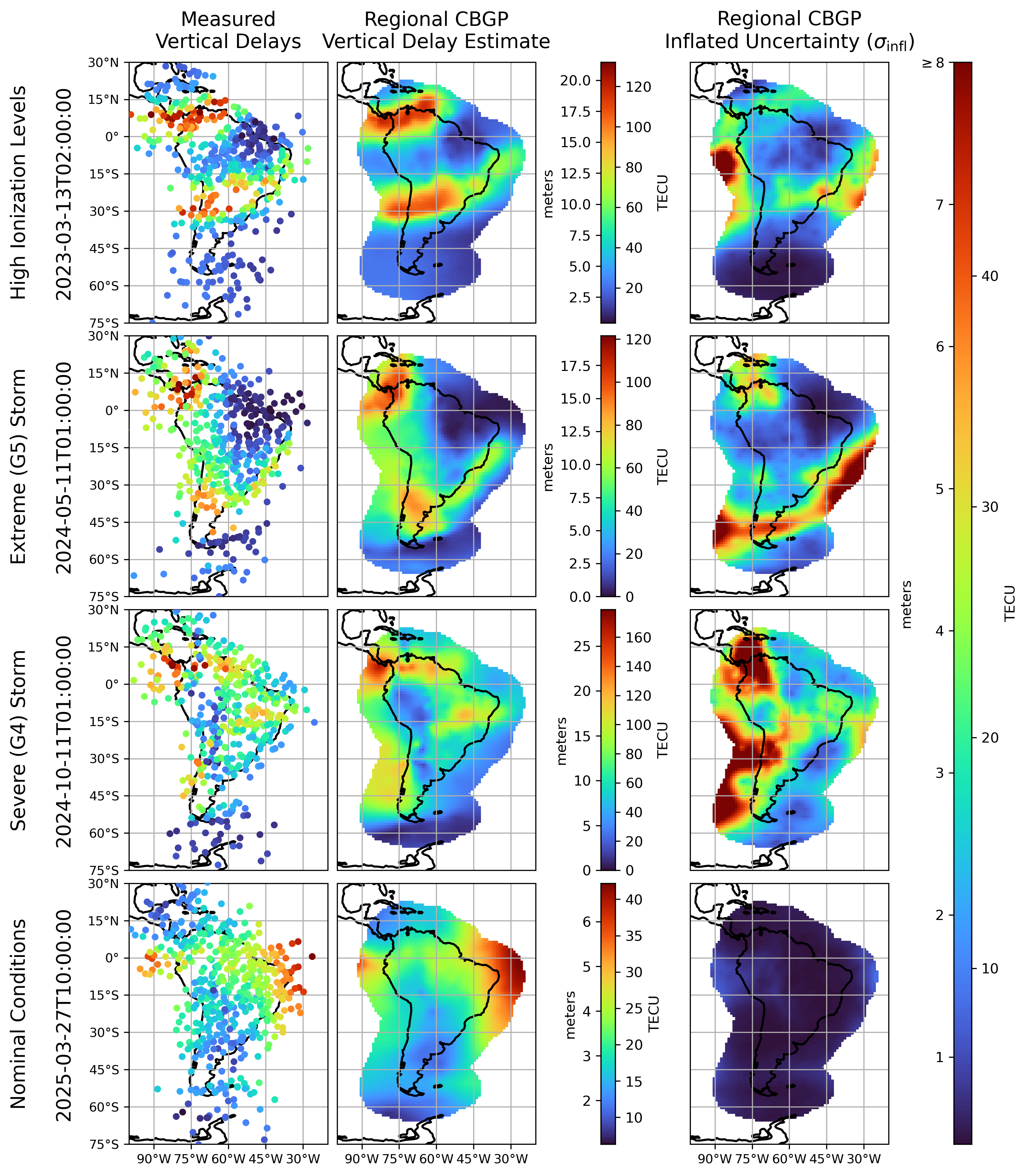}
\caption{Spatial visualizations of ionospheric truth at a single epoch (\textsl{left}) along with regional CBGP posterior means, $\mu$, (\textsl{middle}) and inflated uncertainties, $\sigma_\mathrm{infl}$, (\textsl{right}) for a single epoch from each of the four storms tested. GP posteriors represented on a $1^\circ\times1^\circ$ grid that is different from the IGP set shown in Figure \ref{fig: station_and_IGP_set}.}
\label{fig: ionosphere_solution}
\end{figure}

Figure \ref{fig: ionosphere_solution} provides instantaneous visualizations of ionospheric truth at a single epoch selected from each of the four storms studied along with corresponding posteriors from the regional CBGP model. Vertical delay measurements shown in the left column of the figure, represented both in units of meters and TEC units (TECU), show sample IPPs in the range of approximately 400-550 measurements at each epoch with marker color specifying the measured ionospheric vertical delay. The first three panels in the left column show various space weather disturbances consisting of prominent EIA crests, shown in truth from 13 March 2023, and nonlinear irregularities, observed in truth from the 2024 storms. The bottom left panel shows measurements under nominal conditions from 27 March 2025 during the nighttime, when the ionosphere is much less active than the daytime, with significantly smaller and smoother ionospheric delays than the plots for other disturbed epochs above it. The second column of the figure displays posterior means as predicted by the regional CBGP model on a $1^\circ\times 1^\circ$-resolution grid, capturing the spatial irregularities and nonlinear behavior exhibited in the truth seemingly well. The last column of Figure \ref{fig: ionosphere_solution} shows post-fit inflated uncertainties, represented by $\sigma_\mathrm{infl}$, evaluted at each epoch by the regional CBGP model. Heavily undersampled areas, such as the edges of the hypothetical SBAS service region shown, have exceedingly high uncertainties at or above 8 meters, where a threshold is applied to the plot, during stormy conditions that are significantly inflated by a low effective sample number estimate, $N_\mathrm{eff}$. The uncertainties for the sample nominal conditions from the nighttime ionosphere are much lower than those for disturbed conditions due to notably lower latent signal and observation variation estimates that stem from measurements that are less variable across South America.

Out-of-sample testing is conducted with a deprivation scheme that withholds an entire constellation's worth of measurements at a time for GPR fits. More specifically, GPR fits are performed using only GPS measurements and projected to ionospheric pierce points (IPPs) of deprived Galileo measurements, and, similarly, Galileo measurements would, then, be fit as an in-sample set and then projections would be made at GPS IPPs so that all measurements are tested out-of-sample. Through this construction, all GPR fits are performed using a sampling that would be reflective of an SBAS system that uses a single GNSS constellation for its ionospheric modeling, like WAAS' use of GPS as its lone constellation for ionospheric modeling. For local fit methods, out-of-sample errors are only calculated once for each measurement for fits from its closest IGP. Measurements with IPPs without nearby IGPs from the set shown in Figure \ref{fig: station_and_IGP_set} are excluded from out-of-sample analysis for commensurability of local and regional modeling.

Table \ref{table: iono_results} provides computational results for nearly 28 million out-of-sample ionospheric model errors aggregated over the 20-day storm set for WAAS' locally planar stationary GP, the locally planar CBGP, and the regional CBGP models. First, tests for all three GP models meet the three-nines requirement with over 99.9\% of errors within $3.29\sigma_\mathrm{infl}$. The regional CBGP model's uninflated confidence interval $3.29\sigma$ nearly meets the three-nines requirements on its own as it bounds approximately 99.75\% of the out-of-sample errors. The regional CBGP also tests as the most accurate of the three with the lowest RMSE and MAE values that are at least 10\% lower than its locally planar stationary GP counterpart's while the locally planar CBGP RMSEs and MAEs also improve upon the stationary GP's by at least 8\%. The locally planar CBGP tests yield the lowest NLPD and CRPS values primarily due to its $\sigma$ and $\sigma_\mathrm{infl}$ values that are lower than those for the regional CBGP model.

\begin{table}[htb!]
\begin{center}
\scriptsize
\begin{tabular}{ | c | c | c | c | c | c | c | c | c | }
\hline
\textbf{Model} & \makecell{\textbf{RMSE} \\ \textbf{[m]} } & \makecell{\textbf{MAE} \\ \textbf{[m]} } & \textbf{NLPD} & \makecell{\textbf{CRPS} \\ \textbf{[m]} } & \makecell{\textbf{Within} \\ \textbf{$3.29\sigma$ [\%]} } & \makecell{\textbf{NLPD} \\ \textbf{(infl.)} } & \makecell{\textbf{CRPS} \\ \textbf{(infl.) [m]} } & \makecell{\textbf{Within} \\ \textbf{$3.29\sigma_\mathrm{infl}$ [\%]} } \\ 
\hline
Locally Planar
Stationary GP & 1.01 & 0.592 & 3.26 & 0.480 & 88.07008 & 1.29 & 0.547 & \textbf{99.99133} \\ 
\hline
Locally Planar
CBGP & 0.928 & 0.543 & \textbf{0.885} & \textbf{0.397} & 99.56060 & \textbf{1.09} & \textbf{0.452} & \textbf{99.97186} \\ 
\hline
Regional
CBGP & \textbf{0.901} & \textbf{0.535} & 0.940 & 0.402 & 99.75837 & 1.20 & 0.484 & \textbf{99.98993} \\ 
\hline
\end{tabular}
\end{center}
\caption{Error statistics for out-of-sample ionospheric modeling errors aggregated across 57,600 distinct epochs spanning all 20 storm days and totalling 27,944,482 observations.}
\label{table: iono_results}
\end{table}

\begin{figure}[htb!]
\centering
\includegraphics[width=7in]{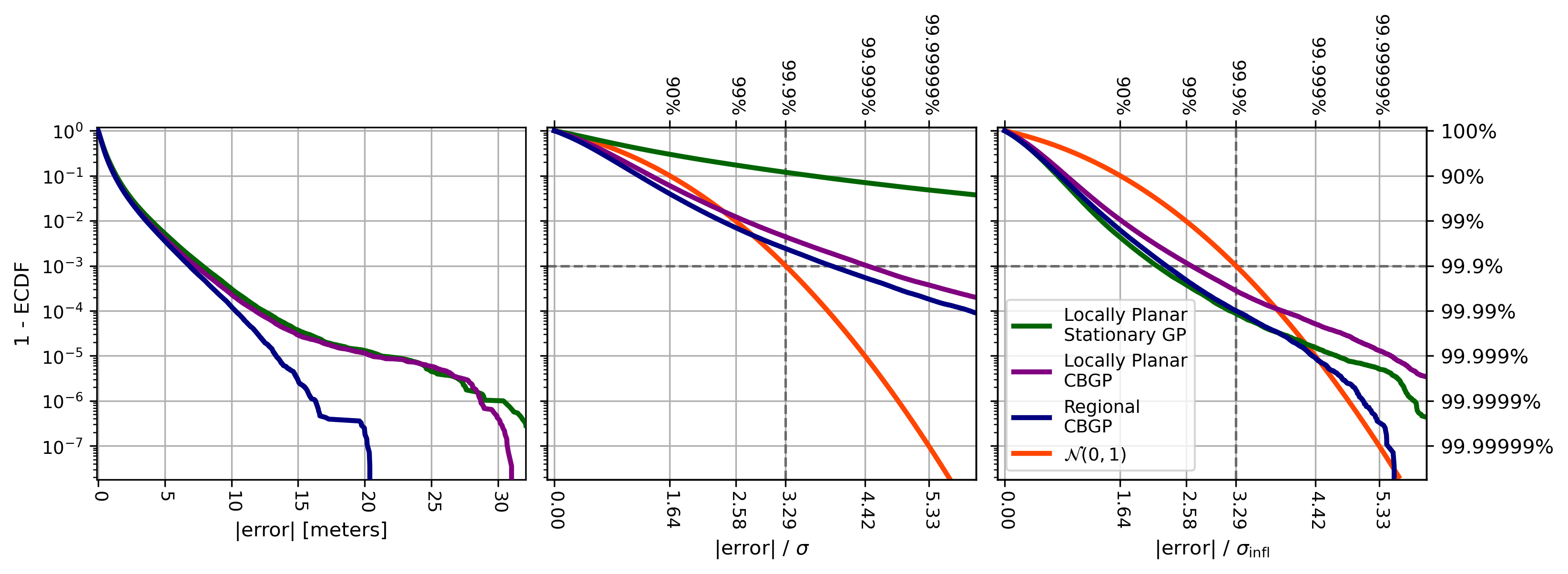}
\caption{Complementary ECDF curves for the magnitudes of out-of-sample errors (\textit{left}), errors standardized by the posterior uncertainty $\sigma$ (\textit{center}), and errors standardized by the inflated posterior uncertainty $\sigma_\mathrm{infl}$ (\textit{right}) aggregated across 57,600 distinct epochs spanning all 20 storm days and totalling 27,944,482 observations.}
\label{fig: iono_ecdf}
 \end{figure}

Complementary ECDF curves for out-of-sample error magntiudes for locally planar stationary GP, the locally planar CBGP, and the regional CBGP models aggregated across the entire ionospheric testing set are shown in Figure \ref{fig: iono_ecdf}. Studying the curves for raw error magnitude distributions, shown in the left figure panel, reveals significantly shorter tails for the regional CBGP in comparison with those for the other two models, suggesting the inappropriateness of planar assumptions in certain contexts that yield out-of-sample errors over 30 meters. Complementary ECDF curves for errors standardized by the posterior uncertainty $\sigma$, as displayed in the center panel of the figure, show approximately 95\% of these CBGP errors bounded by the standard unit normal distribution, shown in orange, while only a small fraction of the locally planar stationary GP errors are bounded. A visualization of each of the three models meeting the three-nines integrity requirement is provided in the right figure panel showing complementary ECDF curves for errors standardized by the inflated posterior uncertainty $\sigma_\mathrm{infl}$ that are below the standard unit normal curve at 3.29.

\begin{figure}[!htb]
 \centering
 \includegraphics[width=4.5in]{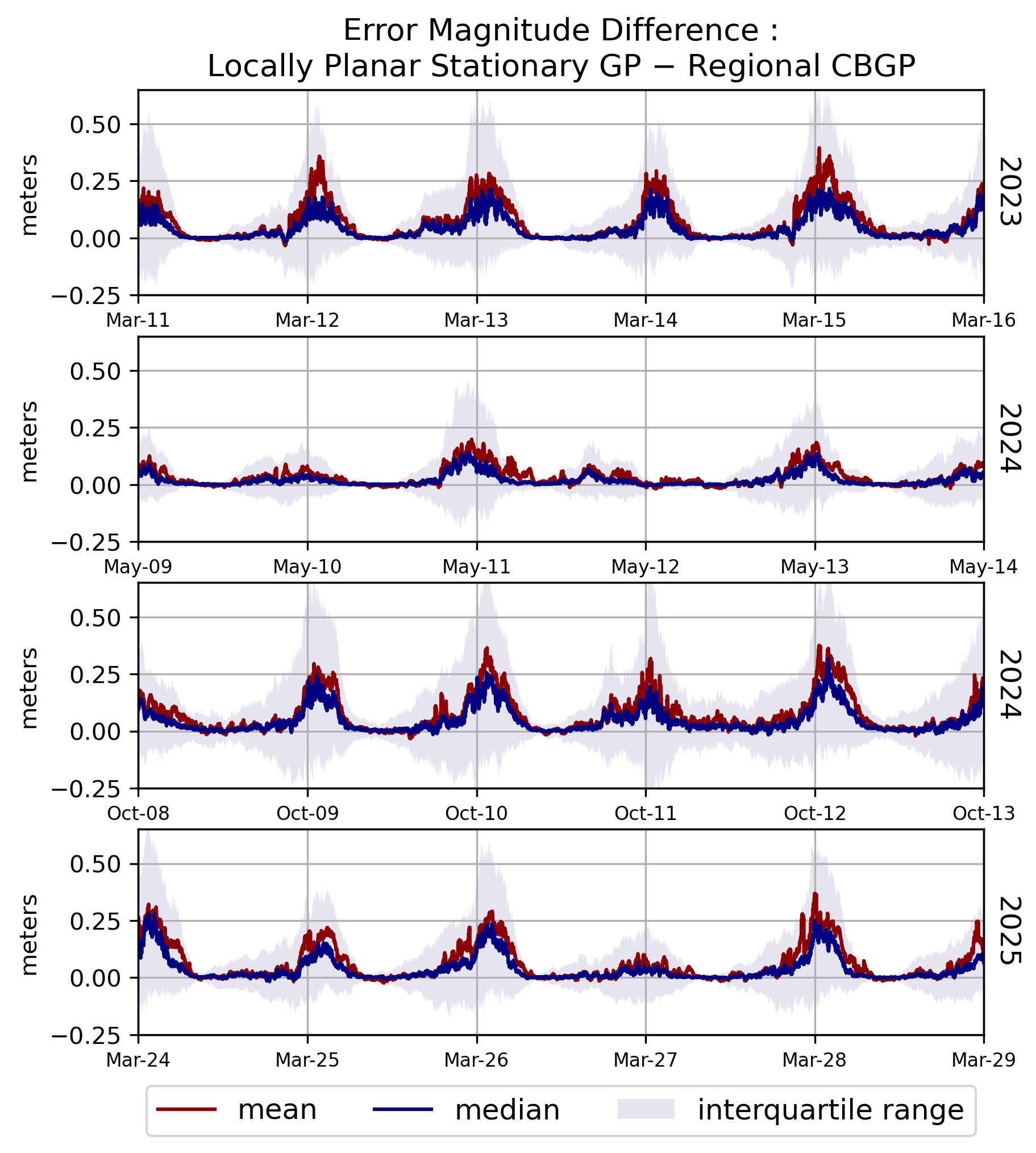}
\caption{Distributions of differences of magntiudes of out-of-sample errors from the locally planar stationary GP and regional CBGP ionospheric models for each of the 57,600 epochs spanning four five-day storms.}
\label{fig: iono_error_over_time}
 \end{figure}

The CBGP model's ionospheric estimation accuracy improvement over WAAS' locally planar stationary GP is diurnally consistent across the entire 20-day testing timeframe. Figure \ref{fig: iono_error_over_time} shows the distribution of differences of out-of-sample raw error magnitudes resulting from the testing of two models at each of the 57,600 epochs tested. In the plots shown, differences of the regional CBGP model errors from the locally planar stationary ones above zero indicate outperformance of the CBGP model from an accuracy standpoint. Daily trends can be observed where model error means and medians at a given epoch are mostly consistent during nighttime and early morning when the ionosphere is significantly less active, occurring during the early-middle of the UTC day. However, during the daytime and early evening, the CBGP model routinely has smaller error magnitudes with rare instances of epochs with underperformance. The mean of the error difference distributions predominantly exceed corresponding medians and 75th percentiles are often further from medians than the 25th percentile during the daytime, indicating a typical skew suggesting heavier positive tails where the CBGP model is more accurate. The regional CBGP model nearly meets or exceeds the accuracy of the locally planar stationary model not only in aggregate, but also for nearly every epoch in our expansive testing set.

\begin{figure}[!htb]
 \centering
 \includegraphics[width=4.5in]{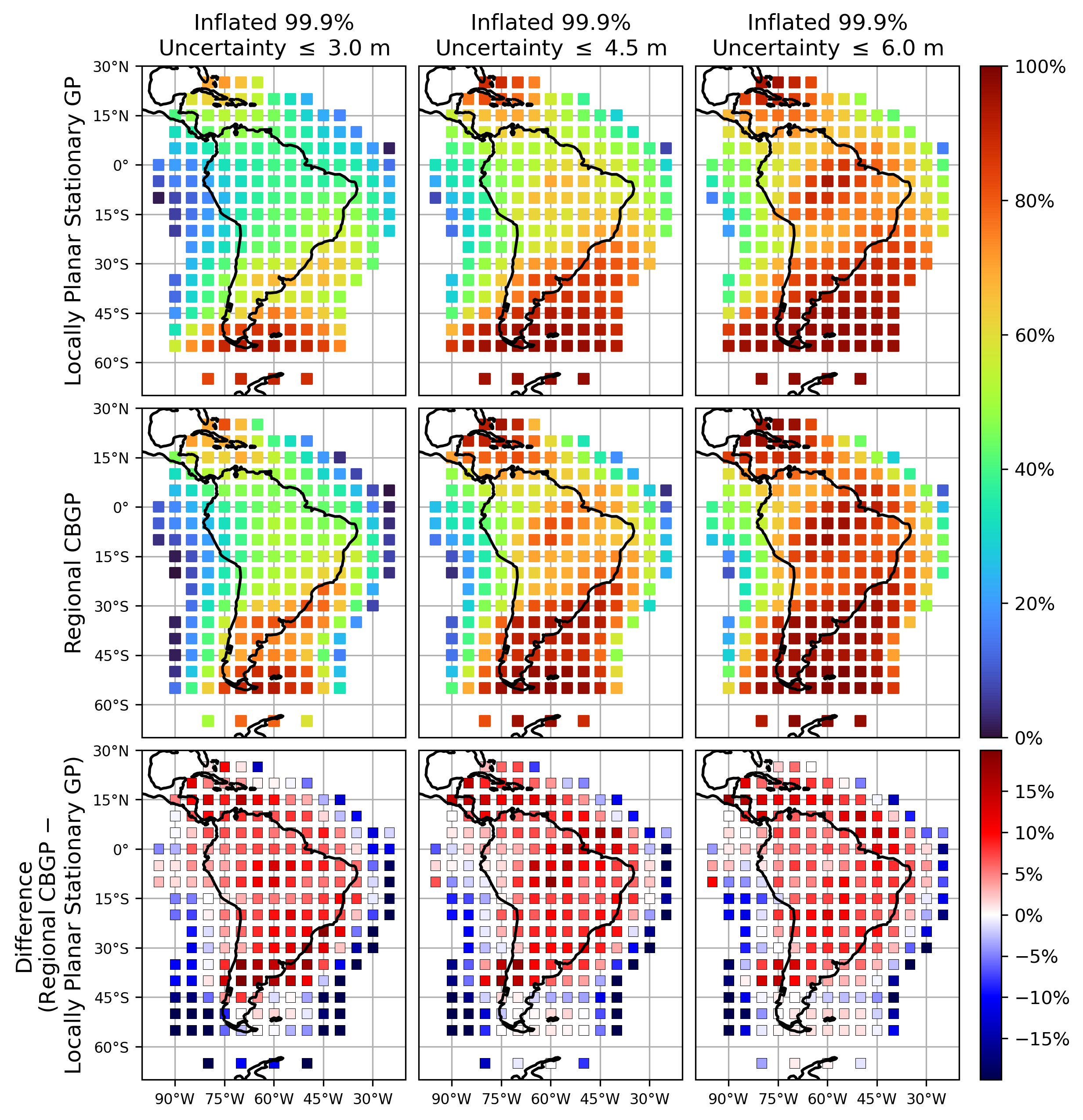}
\caption{Proportions of inflated 99.9\% uncertainties, represented by $3.29\sigma_\mathrm{infl}$, below quantized thresholds of 3.0 (\textit{left}), 4.5 (\textit{center}), and 6.0 meters (\textit{right}) for the locally planar stationary GP (\textit{top}), the regional CBGP (\textit{middle}), and their difference (\textit{bottom}).}
\label{fig: iono_give}
 \end{figure}

In addition to enhanced accuracy, the regional CBGP provides lower uncertainties over South America than the locally planary stationary GP in meeting the three-nines integrity requirement, which would suggest higher availability for a potential SBAS in the region. For a 99.9\% integrity requirement, the MOPS specifies for an SBAS' grid ionospheric vertical errors (GIVEs) to be broadcast to the aviation user as a 3.29-multiple of the ionospheric correction uncertainty, represented by $3.29\sigma_\mathrm{infl}$ \citep{rtca2020}. Lower GIVE values, generally, correspond to smaller positioning uncertainties for the aviation user and, thus, more confident runway approaches for the aviation user. Figure \ref{fig: iono_give} displays the percentage of $3.29\sigma_\mathrm{infl}$ that are bounded by the MOPS-specified GIVE quantization levels of 3.0, 4.5, and 6.0 meters in the left, center, and right columns, respectively, for the locally planar stationary GP in the top row and the regional CBGP in the middle row across the entire 20-day storm set for each IGP without any deprivation of measurements. Over mainland South America, the CBGP's $3.29\sigma_\mathrm{infl}$ values are approximately 5-15\% below their locally planar stationary GP counterparts, as evidenced by the bottom row of the figure that displays the differences between the two. Along the periphery of the IGP set, the inflated uncertanties are higher for the regional CBGP due to the heavy undersampling of the ionosphere there; however, these IGPs may be less important for overall SBAS coverage. Though practical SBAS implementation must account for other factors not considered in this study, computational tests demonstrating higher accuracy and lower uncertainties than WAAS' locally planary stationary GP while still meeting the three-nines integrity requirement are encouraging for fielding a potential SBAS in South America.

\section{Conclusions}
\label{sec: conclusions}

A novel CBGP framework for nonstationary GP modeling is developed in this study to account for spatiotemporal irregularities while cautioning against overfitting or overconfidence. In these models, latent functions for signal and observation variation are estimated by iteratively inflating nominal homogeneous functions through relative error estimation relying on sub-layer GP modeling in partially-whitened domains. The risk of overfitting is mitigated by thresholding these latent functions from above while posterior variances are inflated post-fit to prevent overconfidence. Rigorous testing through data deprivation of both simulated and real-world observations show both accuracy improvement over stationary models and satisfaction of integrity requirements. Notably, all applications of our CBGP modeling framework meet a three-nines integrity standard, revealing that the models are not overly optimistic. Results from benchmark experiments with heavy undersampling and inhomogeneity reveal that CBGP modeling is capable of outperforming existing inhomogeneous GP methods in terms of both accuracy and integrity.

One important aspect of the CBGP modeling in this paper is its robustness, particularly in ionospheric applications that are subject to strict integrity requirements for safety-critical purposes. Ionospheric test cases in the most disturbed conditions both geographically and temporally span a very extensive dataset spanning 20 days of GNSS receiver collections. Nearly all of the spatial GPR fits at each of the 57,600 epochs see our regional CBGP model improve upon the accuracy of a local stationary GP model used by WAAS for ionospheric corrections, suggesting a lack of overfitting that can plague adaptive models. CBGP models also meet the stringent three-nines integrity standard for the ionospheric application, indicating that it is not overconfident. Additionally, the CBGP model is not too computationally expensive to run in real-time due to iterative schemes that can converge relatively quickly with appropriately chosen learning rates. With such accuracy, integrity, and computational efficiency, the CBGP model stands as an excellent candidate for ionospheric modeling in SBAS or Precise Point Positioning-Real-Time Kinematic (PPP-RTK) applications. Adoption of this model would be a stark difference from ionospheric modeling for current SBAS systems that simply fit local spatial neighborhoods using planar assumptions and exclude highly-variable measurements using a pierce point filter line \citep{sparks2022}. Possible practical extensions of our models for ionospheric purposes could be incorporating physics-based GP modeling \citep{yang2021} and characterizing ionospheric threat models \citep{blanch2002} for regional fits using the effective sample number, $N_\mathrm{eff}$.

Even with encouraging computational tests in this study, notable limitations and vulnerabilities accompany the CBGP model in its current form. In Section \ref{sec: models}, seven explicit assumptions for CBGP validity detail the limited appropriateness of the model. Among these assumptions are prior knowledge of the correlation structure, which is usually only speculated or approximated in practical applications. Another notable assumption is the approximate co-location of heightened signal and observation variation for irregulaties, which constrains the scope of potential applications. Any instance in which heteroskedastic noise does not trend spatiotemporally with signal variation may be inappropriate for CBGP modeling. An example of such a scenario may be ionospheric modeling in the polar regions where stochasticity may dominate deterministic elements due to localized plasma irregularities, known as ``patches,'' that can range in size from 10 km to 1,000 km in size \citep{tsunoda1988, zhang2020}. As with any adaptive model, the CBGP may be susceptible to misleading undersampled observations. Such vulnerabilities may be significantly magnified when CBGP models are not well-parameterized or when chosen length scales do not align well with observations.

Though effective in accuracy and integrity testing, the CBGP modeling framework presented in this paper has room for improvement from both a formulation and computational standpoint. Some of the heuristic estimation that attribute partially-whitened variation to signal and observation noise components is the most significant area where improvement can be made and could be potentially replaced with other maximum a posteriori (MAP) estimation-based approaches. Hyperparameters, such as the irregularity length scale for $\rho_\eta$, can be optimally determined instead of prescribed. Also, our methods can be extended in a ``deeper'' fashion that recursively calls the nonstationary framework for multiple layers when applying GP modeling partially-whitened observations instead of using only one extra layer. For large scale applications with large samplings, the inclusion of sparse GP methods could significantly improve computational efficiency and reduce memory costs considerably \citep{quinonerocandela2005, liu2020}. Finally, means to account for correlation length scale heterogeneity, which are assumed to be uniform in this study, may also be incorporated.

The ability of CBGP models to perform well in nominal conditions while also adapt to spatiotemporal irregularities, as evidenced through both simulations and real datasets, make it a strong candidate for ionospheric modeling for SBAS, among other real-time safety-critical potential applications. CBGP's boosting of the prior covariance matrix using relative errors estimated from partially-whitened observations is a novel approach that can prompt the further development of more nonstationary GP methods that improve upon model limitations in this study.


\acks{This research was jointly funded by the United States Federal Aviation Administration (FAA) supporting WAAS and the Australian and New Zealand governments supporting the Southern Positioning Augmentation Network (SouthPAN) initiative. The contents of this paper are the views of the author and do not represent those of the FAA nor the Australian and New Zealand governments. The author appreciates the data provided by IGS, UNAVCO, GFZ Helmholtz Centre Potsdam, and Kyoto University.}


\vskip 0.2in
\bibliography{references.bib}


\appendix


\section{Incorporating Linear Drift Into Gaussian Process Models}
\label{app: incorporating_linear_drift}

For many applications of GPs, linear drift models are used to account for existing trends embedded within observations. Such models assume that observations will deviate from an underlying linear model in a Gaussian fashion, suggesting that linearly detrended observations follow a nearly Gaussian distribution. Ionospheric modeling, one of the notable applications of this study, often assumes a local spatial linearity for fitting measurements captured at a single instant in time \citep{walter2001}.

When assuming an underlying linear drift model, suppose that the observations $y$ deviate from a linear model $y_\mathrm{model}$ with respect to the spatiotemporal state $x$ as follows:
\begin{equation}
\label{eq: y_model}
y_\mathrm{model} = G\beta
\end{equation}
where
\begin{equation}
\label{eq: G_model}
    G = \begin{bmatrix}
        1 & x_{1,1} & x_{1,2} & \cdots & x_{1,n} \\
        1 & x_{2,1} & x_{2,2} & \cdots & x_{2,n} \\
        \vdots & \vdots & \ddots & \vdots \\
        1 & x_{N,1} & x_{N,2} & \cdots & x_{N,n} \\
    \end{bmatrix}
\end{equation}
with $x_{i,k}$ representing the $k$th dimension of the $i$th observation's spatiotemporal state. The $(n+1)$-sized vector $\beta$ represents linear model coefficients. Following \citet{sparks2011a}, the GPR posteriors for expected value and variance are given by
\begin{equation}
\label{eq: gpr_mean_drift}
\mathbb{E}\left[ y^* \right] = w^T y
\end{equation}
\begin{equation}
\label{eq: gpr_var_drift}
\mathrm{Cov}\left[ y^* - w^T y \right] = w^TCw - 2 w^T K^* + K^{**} + R^*
\end{equation}
where
\begin{equation}
\label{eq: w_def_drift}
w= P K^* +Q G^*
\end{equation}
and
\begin{equation}
\label{eq: Q_def_drift}
Q = C^{-1}G(G^TC^{-1}G)^{-1}
\end{equation}
\begin{equation}
\label{eq: P_def_drift}
P = C^{-1} - Q G^TC^{-1}.
\end{equation}
Here, $G^*$ represents the $G$-matrix in Equation \eqref{eq: G_model} evaluated at $x^*$. The whitening matrix for the raw observations with a linear drift, $\Gamma$, is defined by the relationship
\begin{equation}
\label{eq: gamma_with_drift}
\Gamma^T\Gamma = P.
\end{equation}
However, the whitening matrix for the linearly detrended, or ``de-drifted,'' observations would still be defined by Equation \eqref{eq:whitening_definition}.

To de-mean observations using this drift model, the $G$ matrix in Equation \eqref{eq: G_model} can be constructed simply as a column vector of 1s, indicating that the linear model is constant and the observations follow a Gaussian distribution but with a mean that is potentially nonzero. In such scenarios, the drift model serves as a BLUP for de-meaning the observations.

It is worth noting that the methods outlined in Algorithm \ref{alg: cov_infl_alg} do not account for linear drift models, and the algorithm is assumed to be modified appropriately for all such applications in this study.


\section{Parameterizations}
\label{app: params}

Parameters used for all CBGP applications and experiments in Section \ref{sec: results} are provided in Table \ref{table: CBGP_params}. Several parameters from the ``Simulated'' and ``Ionosphere'' columns in the table are sourced from WAAS' operational system, as referenced in \citet{sparks2011a}. The parameters $L_\rho$ and $L_\eta$ represent the kernel length scales for $\rho$ and $\rho_\eta$, respectively.

Parameters used for all stationary GP, hetGPy, and Gibbs-Paciorek model experiments in Section \ref{subsec: benchmark_experiments} are provided in Table \ref{table: other_params}. As is the case for CBGP parameter selection, these parameters are chosen manually to provide results with as high accuracy and integrity as possible. For example, hetGPy motorcycle acceleration experiment with a Gaussian RBF kernel is unable to meet the three-nines integrity standard through our testing, so a Mat\'ern 3/2 kernel is implemented. For Gibbs-Paciorek experiments, stationary model parameters are used to seed initial guesses for latent functions for the length scale, signal variation, and observation variation. In addition, the number of centers used for each experiment are determined roughly as the square root of the sample numbers for the data set after deprivation. 

\begin{table}[htb!]
\begin{center}
\scriptsize
\begin{tabular}{ |c|c|c|c|c| }
\hline
Parameter & Simulated  & \makecell{Motorcycle \\ Acceleration} & \makecell{Meuse \\ River} & Ionosphere \\
\hline
\hline
drift model & None & None & None & \makecell{linear or \\ de-meaning} \\ \hline
\makecell{kernel \\ structure} & \makecell{exponential \\ decay} & \makecell{Gaussian \\ RBF} & \makecell{Gaussian \\ RBF} & \makecell{exponential \\ decay} \\ \hline
$L_\rho$ & 8,000 & 8 ms & 500 m & 8,000 km \\ \hline
$L_\mathrm{\eta}$ & 4,000 & 16 ms & 1,000 m & 4,000 km \\ \hline
$L_\mathrm{eff}$ & 300 & 4 ms & 200 m & 800 km \\ \hline
$\epsilon_\mathrm{eff}$ & 0.25 & 0.25 & 0.25 & 0.25 \\ \hline
$\sigma_\mathrm{signal, 0}$ & $\sqrt{0.91}\approx0.9539$ & 1 g & 0.1 & $\sqrt{0.91}\approx0.9539$ m \\ \hline
$\sigma_\mathrm{obs, 0}$ & 0.3 & 1 g & 0.1 & 0.3 m \\ \hline
$\sigma_\mathrm{s, max}$ & 10 & $\infty$ & 2 & 8 m \\ \hline
$\sigma_\mathrm{o, max}$ & 10 & $\infty$ & 2 & 1.2 m \\ \hline
$z_T$ & 8 & 8 & 8 & 8 \\ \hline
$\kappa_0$ & -0.1257 & -0.1257 & -0.1257 & -0.1257 \\ \hline
$\xi_0$ & 1 & 3 & 3 & 3 \\ \hline
$z_\mathrm{infl}$ & 3.29 & 1.96 & 3.29 & 3.29 \\ \hline
$\gamma_\Psi$ & 4 & 4 & 4 & 4 \\ \hline
$\gamma_\Phi$ & 4 & 4 & 4 & 4 \\ \hline
$\varepsilon$ & 0.05 & 0.05 & 0.05 & 0.05 \\ \hline
\end{tabular}

\end{center}
\caption{Parameters for CBGP applications and experiments in Section \ref{sec: results}.}
\label{table: CBGP_params}
\end{table}

\begin{table}[htb!]
\begin{center}
\scriptsize
\begin{tabular}{ |c|c|c|c| }
\hline
Model & Parameter & \makecell{Motorcycle \\ Acceleration} & \makecell{Meuse \\ River} \\
\hline
\hline
\multirow{6}{*}{\makecell{Gibbs-Paciorek \\ (and Stationary GP)}} & \makecell{kernel \\ structure} & \makecell{Gaussian \\ RBF} & \makecell{Gaussian \\ RBF}   \\ 
\cline{2-4}
& $ L_{\rho,0} $ & 10 ms & 500 m \\
\cline{2-4}
& $\sigma_{\mathrm{signal},0}$ & 10 g & 0.25 \\
\cline{2-4}
& $\sigma_{\mathrm{obs},0}$ & 10 g & 0.25 \\
\cline{2-4}
& \makecell{number of \\ centers} & 8 & 4 $\times$ 4 = 16\\
\cline{2-4}
& \makecell{scaled RBF \\ basis width} & 0.5 & 2 \\
\hline
\multirow{2}{*}{hetGPy} & \makecell{kernel \\ structure} & Mat\'ern 3/2 & \makecell{Gaussian \\ RBF}   \\ 
\cline{2-4}
& $ L_{\rho,0} $ & 10 ms & default \\
\hline
\end{tabular}

\end{center}
\caption{Parameters for stationary, hetGPy, and Gibbs-Paciorek experiments in Section \ref{subsec: benchmark_experiments}.}
\label{table: other_params}
\end{table}

\end{document}